\documentclass[10pt,twocolumn,letterpaper]{article}

\usepackage{cvpr}     % To produce the CAMERA-READY version
\usepackage{graphicx}
\usepackage{amsmath}
\usepackage{amssymb}
\usepackage{booktabs}
\usepackage{enumitem}
\usepackage{multirow}
\usepackage{float}

\usepackage[pagebackref,breaklinks,colorlinks]{hyperref}
\usepackage[dvipsnames]{xcolor}

\usepackage[capitalize]{cleveref}
\crefname{section}{Sec.}{Secs.}
\Crefname{section}{Section}{Sections}
\Crefname{table}{Table}{Tables}
\crefname{table}{Tab.}{Tabs.}

\def\cvprPaperID{10554} % *** Enter the CVPR Paper ID here
\def\confName{CVPR}
\def\confYear{2023}

\usepackage{colortbl}

\newcommand{\xhdr}[1]{\vspace{0em}\noindent{{\bf #1.}}}

\newcommand{\method}{\textsc{DeAR}\xspace}
\newcommand{\pata}{\textsc{Pata}\xspace}
\newcommand{\dataset}{\textsc{Protected Attribute Tag Association}\xspace}

\newcolumntype{a}{>{\columncolor{Gray}}c}
\newcolumntype{b}{>{\columncolor{white}}c}
\newcommand\ccg[1]{\cellcolor{green!35}{#1}} % for cells in second column % and gray colored rows
\newcommand\ccr[1]{\cellcolor{red!25}{#1}} % for cells in second column % and gray colored rows
\usepackage{booktabs} % To thicken table lines
\usepackage{color, colortbl}
\definecolor{LightCyan}{rgb}{0.88,1,1}
\definecolor{Gray}{gray}{0.9}
\definecolor{LightCyan}{rgb}{0.88,1,1}
\newcolumntype{a}{>{\columncolor{Gray}}c}
\newcolumntype{b}{>{\columncolor{white}}c}

\title{DeAR: Debiasing Vision-Language Models with Additive Residuals}

\author{Ashish Seth\thanks{Equal Contribution}\\
IIT Madras, India\\
{\tt\small cs20s030@smail.iitm.ac.in}
\and
Mayur Hemani\footnote[1]~\\
Adobe Inc.\\
{\tt\small mayur@adobe.com}
\and
Chirag Agarwal\\
Adobe Inc.\\
{\tt\small cagarwal@adobe.com}
}

\begin{document}

\makeatletter
\let\@oldmaketitle\@maketitle% Store \@maketitle
\renewcommand{\@maketitle}{\@oldmaketitle% Update \@maketitle to insert...
\vspace{-5pt}
\includegraphics[width=\textwidth]{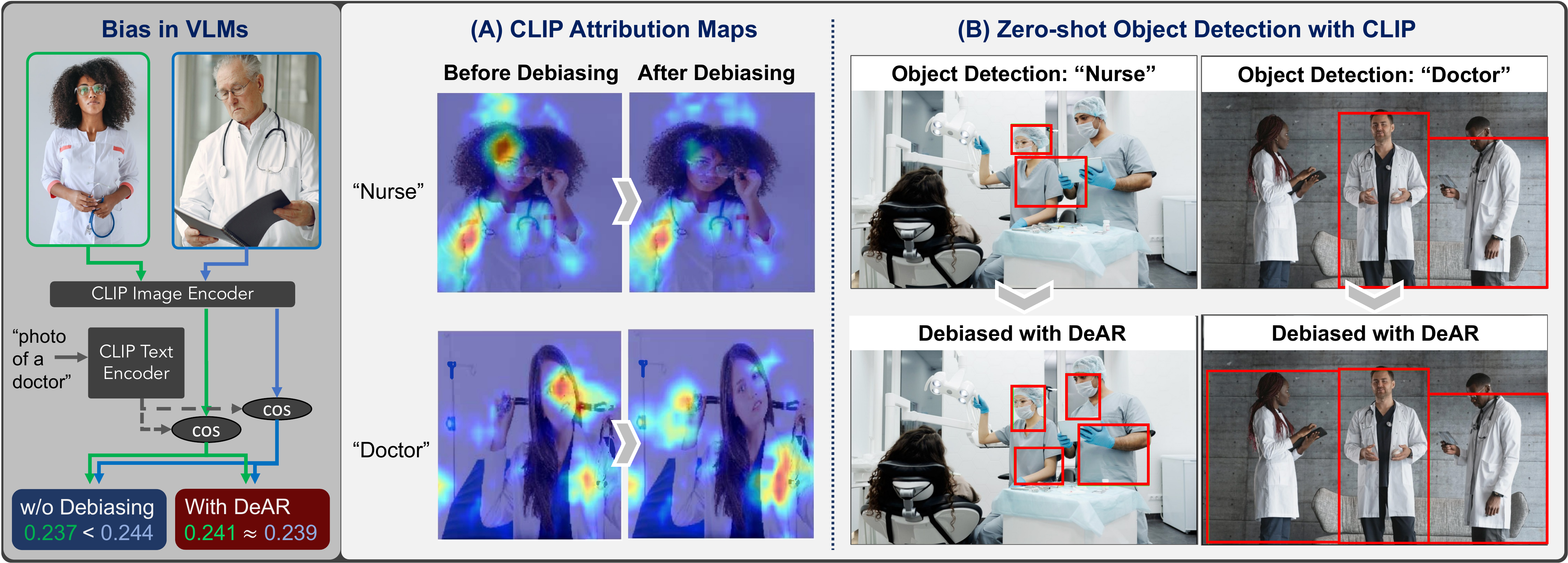}
    \captionof{figure}{We present \method– a framework to de-bias large Vision-Language models (VLM) like CLIP~\cite{radford2021learning}, exhibited in the skewed similarity between specific language concepts and images of people of certain visual characteristics. \textbf{(A)} Attribution maps from the \method-augmented CLIP model indicate how the attribution for a text concept shifts from the person’s facial characteristics to objective cues in the image. \textbf{(B)} Results for zero-shot object detection with CLIP-ODS~\cite{ods} that uses CLIP before and after debiasing show a clear improvement in the fairness of its detection results.}
    \label{fig:teaser}
\bigskip}% ... an image
\makeatother

\maketitle

%%%%%%%%% ABSTRACT
\begin{abstract}
   Large pre-trained vision-language models (VLMs) reduce the time for developing predictive models for various vision-grounded language downstream tasks by providing rich, adaptable image and text representations. However, these models suffer from societal biases owing to the skewed distribution of various identity groups in the training data. These biases manifest as the skewed similarity between the representations for specific text concepts and images of people of different identity groups and, therefore, limit the usefulness of such models in real-world high-stakes applications. In this work, we present \method (Debiasing with Additive Residuals), a novel debiasing method that learns additive residual image representations to offset the original representations, ensuring fair output representations. In doing so, it reduces the ability of the representations to distinguish between the different identity groups. Further, we observe that the current fairness tests are performed on limited face image datasets that fail to indicate why a specific text concept should/should not apply to them. To bridge this gap and better evaluate \method, we introduce the \dataset (\pata) dataset -- a new context-based bias benchmarking dataset for evaluating the fairness of large pre-trained VLMs. Additionally, \pata provides visual context for a diverse human population in different scenarios with both positive and negative connotations. Experimental results for fairness and zero-shot performance preservation using multiple datasets demonstrate the efficacy of our framework.
\end{abstract}

%%%%%%%%% BODY TEXT
\section{Introduction}
\label{sec:intro}
\noindent Deep learning-based vision-language models (VLMs)~\cite{radford2021learning} unify text and visual data into a common representation and reduce the computing cost of training for specific computer vision~\cite{Gu2022OpenvocabularyOD} and visually-grounded linguistic tasks~\cite{shen2021much,cho-etal-2022-fine}. VLMs are trained with a large amount of data with the aim of matching image and text representations for image-caption pairs to capture diverse visual and linguistic concepts. However, VLMs exhibit societal biases manifesting as the skew in the similarity between their representation of certain textual concepts and kinds of images~\cite{berg2022prompt,borchers2022looking,kirk2021bias}. These biases arise from the underlying imbalances in training data~\cite{multimodalDatasetBias, berg2022prompt} and flawed training practices~\cite{biasBeyondBalancedDatasets}. In this work, we present a method to significantly reduce the bias in VLMs by modifying the visual features of the models.

These societal biases in VLMs show up as the selective association (or dissociation) of their representations of human images with specific physical characteristics and their text representations for describing social labels~\cite{Wang2021AreGQ}. For instance, a higher degree of similarity between the representation of the text “doctor” and images of men than that of the women can have trust consequences for models using the representations from these VLMs. To estimate the degree of such biases in VLMs, we compute the cosine similarity between the representations of a set of human images and specific key text phrases and compare their distribution over some associated \textit{protected attributes}. These attributes represent visually discernible characteristics like \textit{gender}, \textit{race}, and \textit{age} common to certain collective identity groups. In this work, we introduce \method, an additive residual-based de-biasing technique that can be augmented with any pre-trained visual-language model with separate visual and language encoders~\cite{radford2021learning, singh2022flava, li2022blip} to improve their fairness.

Our empirical analysis indicates that the protected attribute (PA) information from an image-text pair can be disentangled using their representation by simply learning a linear transformation of the visual representations produced by a VLM and subtracting (adding a negative residual) them from the original representation. We propose to train our framework using two objectives: i) learn a residual representation that when added to the original representation, renders it incapable of predicting different protected attributes, and ii) ensure that this modified representation is as close to the original as possible. We demonstrate that learning \textit{additive residual} enables the de-biasing of pre-trained VLMs using quantitative skew computations on multiple datasets and qualitative evaluations. We also show that the resulting representation retains much of its predictive properties by means of zero-shot evaluation on different downstream tasks.

Recent efforts to mitigate these biases from VLMs, such as by Berg et al.~\cite{berg2022prompt}, use unimodal datasets like the FairFace~\cite{fairface} and UTK~\cite{UTK} datasets. These face-image datasets lack the context necessary to infer the situations in which benign text associations can turn offensive when applied selectively. For instance, if the image of a person drinking water from a bottle is misconstrued as partaking of alcohol and if this association (in terms of image-text similarity) is selective to a specific group of people with some specific visual characteristics, the association may be deemed offensive. To this end, we introduce the \dataset (\pata) dataset to test different associations for VLMs. \pata comprises images of persons in different contexts and their respective text captions with positive and negative connotations. We present our de-biasing results on both the FairFace and the \pata datasets. In summary, the paper makes the following contributions:
\begin{itemize}[nosep]
\item We present the \method framework -- a simple, computationally-efficient, and effective de-biasing method for VLMs that only adapts the image encoder of the VLM by adding a learned residual representation to it. 
\item We introduce the \dataset dataset -- a novel context-based bias evaluation dataset that enables nuanced reporting of biases in VLMs for race, age, and gender protected attributes.
\end{itemize}

\section{Related work}
\label{sec:related}
Our work lies at the intersection of large pre-trained VLMs and the identification, measurement, and mitigation of societal biases in large pre-trained models.

\xhdr{Pre-trained Vision-Language Models} Large VLMs aim to learn general information from a large amount of data and then transfer the knowledge to diverse downstream tasks~\cite{radford2021learning,li2022supervision,li2021align,kim2021vilt,vlmo,singh2022flava,li2020oscar,chen2020uniter}. Recent works~\cite{desai2021virtex,singh2022flava,zellers2021merlot,zhang2021vinvl} leverage contrastive strategies to learn representations for multimodal data. Based on how these models learn these representations, VLMs are broadly categorized into two groups: i) learning image-text representations jointly using transformer-based encoders~\cite{chen2020uniter,li2020oscar,lu202012,li2019visualbert,desai2021virtex,singh2022flava,zellers2021merlot,zhang2021vinvl}, and ii) learning individual unimodal encoders for image and text~\cite{chen2020simple,zbontar2021barlow,chen2021exploring,chen2020improved,he2020momentum,radford2018improving,brown2020language,devlin2018bert}. Models like CLIP~\cite{radford2021learning}, ALIGN~\cite{jia2021scaling}, and BASIC~\cite{pham2021combined} are pre-trained on large datasets collected from the web to bring representations of paired images and text close to each other while distancing them from other pairs. Our work analyzes recently proposed state-of-the-art VLMs for the societal biases they exhibit. Specifically, we look at the behavior of CLIP~\cite{radford2021learning}, FLAVA~\cite{singh2022flava}, and BLIP~\cite{li2022blip} and propose to alleviate bias from their visual representations.

\xhdr{Fairness Techniques} Prior work in language~\cite{borchers2022looking,guo2021detecting,kirk2021bias}, computer vision~\cite{wang2022revise,gebru2021datasheets}, and graphs~\cite{agarwal2021towards,kang2021fair,wang2022unbiased,ma2022learning} has primarily focused on debiasing models trained on unimodal data and are limited in scope as they only investigate gender bias, racial bias, or their intersections. In particular, their bias mitigation techniques can be broadly categorized into i) \textit{pre-processing}, which modifies individual input features and labels~\cite{calmon2017optimized}, modifies the weights of the training samples~\cite{kamiran2012data}, or obfuscates protected attribute information during the training process~\cite{zemel2013learning}; ii) \textit{in-processing}, which uses adversarial techniques to maximize accuracy and reduce bias for a given protected attribute~\cite{zhang2018mitigating}, data augmentation~\cite{agarwal2021towards} or adding a bias-aware regularization term to the training objectives~\cite{kamishima2012fairness}, and iii) \textit{post-processing}, which changes the output predictions from predictive models to make them fairer~\cite{kamiran2012decision,pleiss2017fairness,hardt2016equality}. It is non-trivial to apply these techniques to pre-trained VLMs because the training requires a large annotated dataset and immense computing resources. Mitigating bias in large pre-trained VLMs is a nascent research direction. Wang et al.~\cite{wang2021gender} propose to remove the dimensions in CLIP embeddings most associated with gender bias, while Berg et al.~\cite{berg2022prompt} use adversarial fine-tuning with a corpus of face images and arrays of text prompts to mitigate bias. The former cannot achieve joint mitigation of bias for multiple protected attributes, and the latter method modifies the original model changing its accuracy on zero-shot tasks significantly. Our proposed \method framework addresses both issues by training a lightweight residual representation over the visual representations of VLMs, modeling the joint bias with respect to multiple protected attributes (gender, race, and age), and ensuring the similarity of the modified representation to the original. 

\looseness=-1
\xhdr{Fairness Benchmarks} Previous work on debiasing VLMs~\cite{wang2021gender, berg2022prompt} exclusively focuses on face-image datasets, \ie, FairFace~\cite{fairface} and UTKFace~\cite{UTK} datasets. While the FairFace dataset has 10,954 test images consisting of faces of 10 age
groups, seven race groups, and the binary genders, the UTKFace dataset has 23,708 cropped test images in 104 different age values (1-105), five race groups, and the binary genders. Neither datasets have the pixels or annotations to provide the context of the person in an image. We collect a dataset of 4934 test images organized into different contextual scenes
and prepare positive and negative captions for each context, providing a more nuanced view of a VLM’s fairness.

Next, we discuss how we achieve this accuracy-preserving joint mitigation of bias from multiple VLMs.

\section{Preliminaries}
\label{sec:prelims}
We formally define the notion of fairness for VLMs, stipulate the fairness problem, and motivate the use of an additive residual for bias mitigation in VLMs. 

\xhdr{Visually Discernible Protected Attributes (VDPAs)} A \textit{protected attribute} (PA) is defined as any characteristic of a group of people associated with their collective identity. A subset of these attributes are visually discernible, \ie, it's possible to label an image of a person belonging to a particular identity group with the corresponding protected label (PL). Our study includes the i) the binary {\small\texttt{gender}} of a person, ii) the ethno-racial characteristics (skin color, face shape, hair color, etc.) captured as {\small\texttt{race}}, and iii) the {\small\texttt{age}} of a person. For simplicity, we associate a single ground-truth VDPA label for an image $\mathbf{I}_{i}$, which has a gender label $y^{g}_{i}$, a race label $y^{r}_{i}$ and an age label $y^{a}_{i}$. We denote a protected attribute as $P(\mathbf{I})$ with the corresponding labels as a set \{${p_i}$\}.

\xhdr{Abstract Model of Vision-Language Models} An abstract VLM consists of two key components: i) an image encoder $E_i$ from the space of a 2D, 3-channel image to a latent space in $\mathbb{R}^d$ and ii) a text encoder $E_t$ from the space of text tokens to a similar latent space in $\mathbb{R}^d$. For an image $\mathbf{I}$ and text token sequence $\mathbf{T}$, the output of the image and text encoders are denoted as $E_i(\mathbf{I})$ and $E_t(\mathbf{T})$. Finally, a fusion function combines the encoded representations from $E_i$ and $E_t$ into a common representation, which we disregard because of the manner in which they compute similarity.

\xhdr{Similarity of Image and Text Representations} Most large pre-trained VLMs~\cite{radford2018improving,li2022blip,singh2022flava} learn image and text representations that are in a similar latent space. We compute the similarity between an image ($\mathbf{I}$) and a text sequence ($\mathbf{T}$) using the cosine similarity between their representations:
\begin{align}
    \begin{split}
        \text{Sim}_E(\mathbf{I}, \mathbf{T}) &= \frac{E_i(\mathbf{I})\cdot E_t(\mathbf{T})^\top}{||E_i(\mathbf{I})||~ ||E_t(\mathbf{T})||},
    \end{split}
    \label{eqn:similarity}
\end{align}

Formally, the VLM encoders are fair for a protected attribute if the similarity between an image $\mathbf{I}$ and a text sequence $\mathbf{T}$ is independent of the protected label for $\mathbf{I}$. It is formalized by measuring the skew of the similarity scores of $\mathbf{T}$ with a set of images, across their protected labels.\footnote{The formulation used here is adapted from \cite{maxSkew}.}

\xhdr{Fairness and Fair Encoders} We denote the probability of randomly selecting an image $\mathbf{I}$ from a set of images $\mathcal{I}$ such that it has a protected attribute label $P(\mathbf{I})=p_i$ with $f_i$. Next, we define a matching function $M_\epsilon(\mathbf{I}, \mathbf{T})$ which is 1 if $\text{Sim}_E(\mathbf{I}, \mathbf{T}) \geq \epsilon$, and 0 otherwise. 

We define $\mathcal{I}^m$ as the subset of $\mathcal{I}$ for which $M_\epsilon(\mathbf{I}, \mathbf{T})$ is 1 for each $\mathbf{I}\in\mathcal{I}^m$ and the fraction of these images that also have the protected label $p_i$ is denoted by $f^m_i$.

The ratio of $f^m_i$ and $f_i$ is considered as the skew in the distribution of images with the protected label $p_i$ imposed by the model ($E_i, E_t$) by virtue of the similarity between its output representations. The \textit{Skew} measure for the $i^{th}$ protected label, with respect to the text sequence and image set, is then computed as:
\begin{align}
    \begin{split}
        \text{Skew}_i(\mathcal{I}, \mathbf{T}) &= \log (f^m_i / f_i),
    \end{split}
    \label{eqn:skew}
\end{align}
where the score is zero when $f^m_i = f_i$, \ie, the matching does not depend on the value of the protected attribute.

The goal of learning a fair encoding then consists of training encoders ($E_i^{*}$, $E_t^{*}$) such that $\text{Skew}_i(\mathcal{I}, \mathbf{T}) \approx 0$ for all text sequences $\mathbf{T}$, protected attributes $i$, and for any set of images $\mathcal{I}$. We posit that a model devoid of any information useful for distinguishing between different PLs for a PA is fair because encoding from such a model is independent of that particular PA. We submit that learning such a model with respect to a VLM is possible by adapting the visual encoder of the VLM. This is because the encoding from it captures rich information in a compact form, and it is easier to find training data to adapt the image encoder than it is for the text encoder. We posit that adapting just the image encoder can produce a fair model.  We corroborate this position next with empirical evidence.

\xhdr{Disentangling Protected Attribute Information} To motivate the design of \method, we present empirical evidence for the bias problem using the {\small\texttt{gender}} attribute. We show that $E_i(\mathbf{I})$ can be split into two additive components - i) $\phi(\mathbf{I})$ that represents the gender information for $\mathbf{I}$ and ii) a protected-attribute (PA) free representation $\overline{\phi}(\mathbf{I})$ that cannot be used to identify the gender of the person in the image:
\begin{align}
    \begin{split}
        E_i(\mathbf{I}) &= \overline{\phi}(\mathbf{I}) + \phi(\mathbf{I})
    \end{split}
\end{align}

For the protected labels {\small{\texttt{male(m)} and \texttt{female(f)}}}, we get:
\begin{eqnarray}
    E_i(\mathbf{I}_{f}) = \overline{\phi}(\mathbf{I}_{f}) + \phi(\mathbf{I}_{f}) \\
    E_i(\mathbf{I}_{m}) = \overline{\phi}(\mathbf{I}_{m}) + \phi(\mathbf{I}_{m}) \\
    \Rightarrow E_i(\mathbf{I}_{f}) - E_i(\mathbf{I}_{m}) = \phi(\mathbf{I}_{f}) - \phi(\mathbf{I}_{m}),
    \label{eqn:diffvec}
\end{eqnarray}
where $\overline{\phi}(\mathbf{I}_{f}){=}\overline{\phi}(\mathbf{I}_{m})$ as both contain the PA free representation based on the context in the image. Further, we hypothesize that $\phi(\mathbf{I})$ can be adequately captured for a VLM by the corresponding text encoding for the equivalent textual description ``photo of a man/woman" ($t_{m}$/$t_{f}$) using a linear transformation. This is expressed as:
\begin{align}
    \begin{split}
      \phi(\mathbf{I}_f) &= \mathbf{A}\cdot 
      E_t(t_{f}) + \mathbf{B}\\
      \phi(\mathbf{I}_m) &= \mathbf{A}\cdot E_t(t_{m}) + \mathbf{B},
    \end{split}
    \label{eqn:linear}
\end{align} 
where $\mathbf{A} \in \mathbb{R}^{d \times d}$ is a transformation matrix and $\mathbf{B} \in \mathbb{R}^{d}$ is a vector of the same dimensions as the representations. Combining Eqns.~\ref{eqn:diffvec} and \ref{eqn:linear}, we get:
\begin{align}
    \begin{split}
    E_i(\mathbf{I}_{f}) - E_i(\mathbf{I}_{m}) = \mathbf{A}\cdot &E_t(t_{f}) - \mathbf{A}\cdot E_t(t_{m})\\
    \Rightarrow \mathbf{K}[E_i(\mathbf{I}_{f}) - E_i(\mathbf{I}_{m})] = E_t&(t_{f}) - E_t(t_{m}),
    \end{split}
    \label{eqn:diffequiv}
\end{align}
where $\mathbf{K}=\mathbf{A}^{-1}$. If we can find evidence that such a linear transformation $\mathbf{K}$ of the difference of image representation exists that matches the difference between the encoding of the corresponding text concepts, then we can learn a linear function that disentangles $\phi(\mathbf{I})$ from $E_i(\mathbf{I})$. 

\looseness=-1
\noindent \textit{Empirical evidence:} We train a regression model on 10000 random pairs of images of women and men in a professional workplace setting (a subset of the FairFace~\cite{fairface} dataset), to learn a linear function from the difference of the CLIP image representations to the difference of the CLIP text representations for the ``photo of a woman'' and ``photo of a man'' phrases. We found the model to have a bounded relative mean-squared error (${<}10^{-6}$) over 100 repetitions of the experiment for unseen pairs, which demonstrates that a piece-wise linear function can possibly be trained to approximate $\phi(\mathbf{I})$. Next, we describe how our \method framework trains to approximate $\phi(\mathbf{I})$ for multiple protected attributes, and uses it to remove protected attribute information from $E_i(\mathbf{I})$. 

\section{Our Framework: \method}
\label{sec:method}
\begin{figure*}
    \centering
    \includegraphics[width=\textwidth]{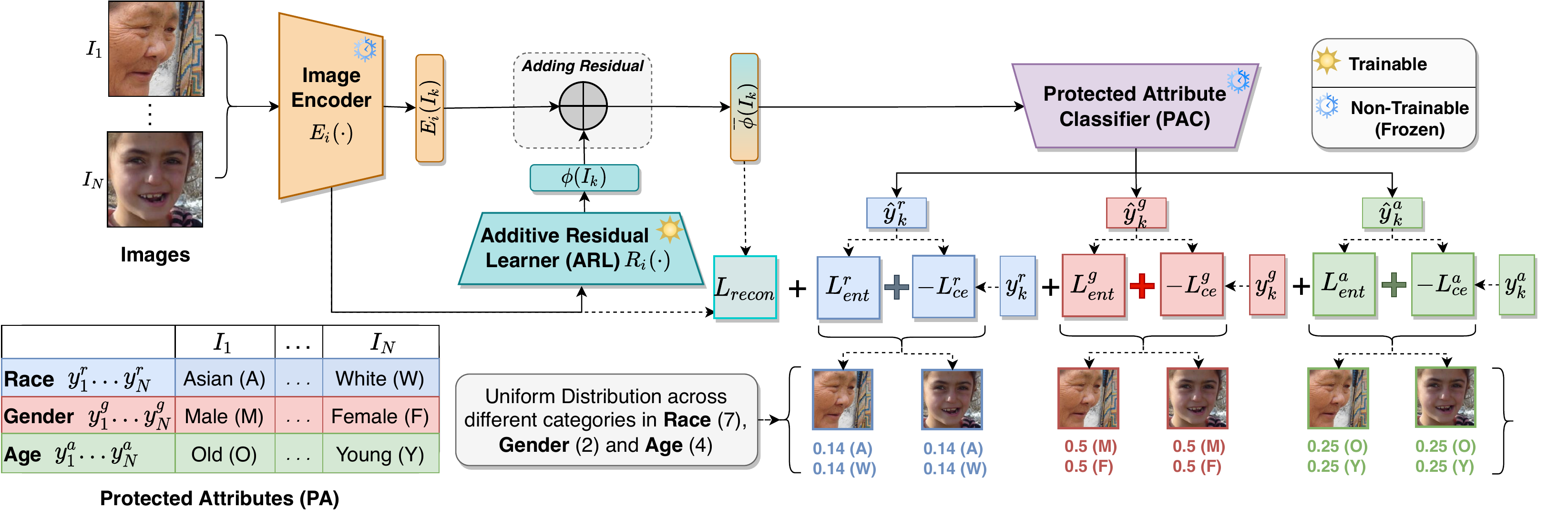}
    \caption{Our \method framework can learn fair representations for pre-trained Vision-Language models. 
    \method is a debiasing framework that takes image representations $E_i(I_k)$ as input and returns a residual representation $\phi(I_k)$. Next, it adds $\phi(I_k)$ with $E_i(I_k)$ to get debiased representation $\overline{\phi}(I_k)$ that is fed to a protected attribute classifier to get prediction for multiple protected attributes (race $\hat{y}^r_k$, gender $\hat{y}^g_k$, and age $\hat{y}^a_k$). \method jointly minimizes the negative of Cross Entropy loss for race ($L^r_{ce}$), gender ($L^g_{ce}$), and age ($L^a_{ce}$) attributes for debiasing across multiple protected attributes and a reconstruction loss $L_{recon}$ to retain the zero-shot performance of the underlying VLM.
    }
    \label{fig:dear-framework}
\end{figure*}

Our framework comprises of two key components: i) an Additive Residual Learner (ARL) component that learns to disentangle the protected attribute information $\phi(\mathbf{I})$ from the image representation produced by the visual encoder of a given VLM and ii) the protected attribute misclassification objective designed to train the ARL component. We first discuss the ARL component and then present details of the training network and objectives.

\subsection{Additive Residual Learning}
\label{sec:arl}
As discussed in Sec.~\ref{sec:prelims}, we propose to train a network $R(\cdot)$ that outputs a representation $\phi(\mathbf{I})$ so that we can compute the protected-attribute free representation $\overline{\phi}(\mathbf{I})$:
\begin{align}
    \overline{\phi}(\mathbf{I}) &= E_i(\mathbf{I}) - R(E_i(\mathbf{I}))
\end{align}
For brevity, we denote $R(E_i(\mathbf{I}))$ as $R(\mathbf{I})$. Note that the network $R: \mathbb{R}^d\rightarrow \mathbb{R}^d$ is a single linear transformation, where $d$ is the number of dimensions of the image representation produced by $E_i$. This is followed by a function $\sigma_{\text{act}}(\cdot)$ that represents the combination of activation and regularization layers (dropout, etc.) as used by the last layer of the image encoder $E_i$. If the last layer does not use any activation function, $\sigma_{\text{act}}$ is an identity transformation, \ie,
\begin{align}
    \begin{split}
        \overline{\phi}(\mathbf{I}) &= E_i(\mathbf{I}) + \sigma_{\text{act}}( R(\mathbf{I})),
    \end{split}
\end{align}
where $\overline{\phi}(\mathbf{I})$ represents the protected-attribute free representation for image $\mathbf{I}$. Note that the reason for transforming the last layer activation and regularization mechanisms from the image encoder is to align the embedding space, where $\overline{\phi}(\mathbf{I})$ maps to the original encoding $E_i(\mathbf{I})$ so that the image-text equivalence still holds. The resulting representation has two primary goals: i) it matches the original representation so that the zero-shot accuracy of the model can be retained, \ie, $\overline{\phi}(\mathbf{I})\approx E_i(\mathbf{I})$, and ii) it cannot be used to distinguish the protected label for the image with respect to different protected attribute. The above goals are achieved through different training objectives, which we describe next.

\subsection{Training ARL with Protected Attribute Misclassification}
\label{sec:paco}
To train the ARL to produce representations that cannot be used to identify the protected labels for an image, we pre-train an adversary -- the Protected Attribute Classifier (PAC) network and use its classification loss as an objective to be maximized for training the ARL. Unlike previous works in bias mitigation that jointly trains an adversarial network, we separately train the PAC. This is possible because the PAC can be trained with the underlying VLM representations and be used to supply gradients as the representations change to their protected-attribute-free form during training. 

The PAC module uses $\ell_{2}$ normalized $d$-dimensional representation from the VLM image encoder $E_i$ as input and generates multiple sets of logits for individual protected attributes. It consists of a single linear layer ($d\times256)$ with ReLU activation, followed by multiple linear classification projection heads ($256\times128$) with ReLU, and a linear layer to produce logits with output sizes 7 (for race), 4 (for age) and 2 (for gender), where the logit sizes are determined by the FairFace~\cite{fairface} training dataset. A unified architecture like \method helps in jointly modeling multiple protected attributes associated with an image. The network is trained on the combined cross-entropy loss using the ground-truth protected labels (race, age, and gender) for each image:

\begin{multline}
\label{eqn:info}
\mathcal{L}_{\text{PAC}} = \mathcal{L}^r_{\text{ce}}(y^{r},\hat{y}^r) + \mathcal{L}^g_{\text{ce}}(y^g,\hat{y}^g) + \mathcal{L}^a_{\text{ce}}(y^a,\hat{y}^a),
\end{multline}
where $\hat{y}^r$, $\hat{y}^g$ and $\hat{y}^a$ are the predictions, and $y^r$, $y^g$ and $y^a$ are the ground-truth labels for the protected attributes race, gender, and age respectively. 

\xhdr{Training objectives for the ARL module} The VLM parameters are frozen while the ARL module is trained using three objectives: i) minimizing the reconstruction loss $L_{\text{recon}}$, which is the mean $\ell_{2}$ loss between the unmodified image representations $E_i(\mathbf{I})$ and the modified protected attribute information-free $\overline{\phi}(\mathbf{I})$, ii) the $L_{\text{ent}}^{\{r,g,a\}}$ to minimize the maximum soft-max probability for each protected attribute classifier head, and iii) $L_{ce}^{r,g,a}$, to maximize the misclassification of the protected label by the PAC module.

\begin{align}
    \begin{split}
L_{\text{ARL}} &= w_{\text{recon}}\cdot L_{\text{recon}} + w_{\text{ent}}\cdot (L_{\text{ent}}^r + L_{\text{ent}}^g + L_{\text{ent}}^a) \\
& -(w_{\text{ce}}^r\cdot L^r_{\text{ce}} + w_{\text{ce}}^a\cdot L^a_{\text{ce}} + w_{\text{ce}}^g\cdot L^g_{\text{ce}}),
    \end{split}
    \label{eqn:arl_objective}
\end{align}
 where $L_{\text{recon}} {=} \|E_i(\mathbf{I}) {-} \overline{\phi}(\mathbf{I})\|_2$, $L_{\text{ent}}^x {=} \max (\sigma_{\text{softmax}}(\hat{y_{x}}))$ (for PA $x$), $\sigma_{\text{softmax}}(x) = \text{Softmax}(x)$, and the $w_i$'s are the regularization weights that assist the training of \method across multiple protected attributes.

The negative cross-entropy losses aim to maximize the cross-entropy loss for the PAC. However, when used alone, this loss results in the protected label predictions getting flipped (from one label to another) rather than reducing the decision confidence of the PAC. The $L_{\text{ent}}$ ensures that the PAC module produces high-entropy logits from the modified image representation $\overline{\phi}(\mathbf{I})$ so that it cannot be easily used for predicting the protected attributes with high confidence. Figure~\ref{fig:dear-framework} describes the operation of the \method framework in detail. The success of the network is evaluated by the degree to which the i) bias in the model is mitigated and ii) zero-shot performance of the model is retained.

\section{Experiments}
\label{sec:evaluation}
Here, we describe the datasets, the metrics, and the implementation parameters for our experiments.

\xhdr{Datasets}
\label{sec:dataset}
We use two datasets each consisting of a set of images and their ground truth (protected labels) to analyze the bias in the output representations of VLMs. Further, each image in the datasets is tested  against a set of curated text captions that represent different desirable (positive) and undesirable (negative) connotations.

\xhdr{\textit{Images and Protected Labels}} We obtain images and labels from the following:
% \begin{enumerate}[nosep]
% \item 
i) the \textit{FairFace dataset} \cite{fairface} consists of a set of cropped face images of different persons. It has 86,700 training images and 10,954 validations images, annotated with the binary gender (male and female), ethnic-racial classes (White, Black, Indian, East Asian, Southeast Asian, Middle Eastern, and Latino), and multiple age brackets, which we bin into four categories (\textit{child} $<20$, \textit{young} $<40$, \textit{middle-aged} $<60$, and \textit{senior} $>60$), and
% \item 
ii) the \textit{\dataset} (\pata) dataset consists of 4,934 public images crawled from the web, organized in 24 scenes each with between 100 and 400 images. These scenes represent situations in which humans of specific identity groups may have biased notions associated with them. The images are further annotated with the binary gender, ethnic-racial labels (White, Black, Indian, East Asian, and Latino-Hispanic), and two age group labels (young and old adults).
% \end{enumerate}

\xhdr{\textit{Text Captions}} Each scene in the \pata dataset is associated with the positive caption set consisting of desirable descriptions of a person, designed specifically to elicit a biased response from the models. For instance, for the {\small\texttt{race}} protected attribute (PA), the caption ``photo of a law-abiding citizen'', and for the {\small\texttt{gender}} protected attribute, the caption ``photo of a smart engineer'' must both result in uniformly distributed similarities between the images and the text. The negative captions similarly consist of untoward and offensive text in the context of each protected attribute. For the \pata dataset, there are 78 (unique) positive captions, and 88 negative captions, with an average of 1.08 and 1.2 per (scene, PA) pairs, respectively. For the FairFace dataset, we consider all images to belong to a single scene and use 16 positive and negative captions for each.

\xhdr{\textit{Advantages of the \pata dataset}} Images in the FairFace dataset only have cropped faces and, therefore, do not contain any context. Hence, any association between the text captions and the face images is incidental at best because the neural network does not have any signal to form the association. The \pata dataset bridges this gap and thus enables a more nuanced evaluation of the bias presented by a VLM. More details about the dataset are included in the Appendix~\ref{app:dataset} of the supplemental material.

\xhdr{\textit{Zero-shot Evaluation datasets}} We consider the CIFAR-10, CIFAR-100~\cite{cifar}, FER2013~\cite{fer2013}, and ImageNet~\cite{imagenet} datasets for zero-shot image classification task using debiased VLM representations. Further, we consider the UCF-101~\cite{ucf101}, Kinetics-700~\cite{kinetics}, and RareAct~\cite{rareact} datasets for zero-shot video classification.

\xhdr{Evaluation Metrics}
\label{subsec:metrics}
Our de-biasing framework is evaluated on two key criteria: i) the degree to which it improves the fairness of a given VLM, which is measured using the similarity between person images with different protected labels and a set of text captions, and ii) the degree to which it retains the zero-shot performance of the underlying VLM on various tasks, including classification on human and non-human related datasets.

\xhdr{\textit{Fairness Metrics}} The degree of fairness exhibited by a model is measured in two respects:

\noindent i) the overall shift in similarity characteristics of the image and text features towards fairness measured using the mean \textit{MaxSkew} and the mean \textit{MinSkew} scores. The former indicates the selective association of the model in favor of any protected label and the latter indicates the degree of selective dissociation against any other protected label. These are computed for a set of images $\mathbf{I}$, a set of captions $\mathbf{T}$, and a protected attribute P with labels {$p_i$} as:
\begin{align}
    \begin{split}
        \psi_{max}(\mathbf{I}, \mathbf{T}, P) &= \text{Mean}_{t\in \mathbf{T}}[\max_{i}(\text{Skew}_{i}(\mathbf{I}, t))] \\
        \psi_{min}(\mathbf{I}, \mathbf{T}, P) &= -\text{Mean}_{t\in \mathbf{T}}[\min_{i}(\text{Skew}_{i}(\mathbf{I}, t))],
    \end{split}
\end{align}
where $\text{Skew}_{i}(\cdot)$ is computed using Equation~\ref{eqn:skew}.

\noindent ii) the ordered measurement of the similarity distribution across the protected labels of each protected attribute, which is indicated using the \textit{MaxSkew@K} and \textit{MinSkew@K} scores as stipulated in Geyik et al.~\cite{maxSkew} and used for fairness analysis. This is a ranking-based measurement that only considers the $k$ most similar images to each text caption, and then computes the mean Max- and Min-Skew scores. To add more nuance to the evaluation, we compute the above metrics for both positive and negative captions for the person in the image, which clearly indicates the stakes for the model in selective association or dissociation for a particular identity group.

\xhdr{\textit{Zero-shot performance metrics}} We evaluate the zero-shot performance of \method-augmented and vanilla VLMs using standard classification accuracy metrics for the image-classification tasks. For video action recognition tasks, we use the top-1 accuracy, the averaged accuracy measures and variants of the mean weighted average precision metrics as described in Miech et al.~\cite{miech20rareact}. We report the drop in each metric to indicate how much the debiasing process affects the zero-shot performance of the underlying VLM. In addition, we discuss the effect of \method framework on the task of zero-shot open-vocabulary object detection~\cite{ods}.

\xhdr{Implementation Details}
\label{sec:training}
Both PAC and the ARL modules are individually trained for each VLM using the training subset of the FairFace dataset, which consists of 86,700 images with annotations for the race, gender, and age attributes. The PAC is trained on its objectives (Equation~\ref{eqn:info}) for 10 epochs, whereas the ARL is trained for 30 epochs with early stopping based on validation loss saturation. See Appendix~\ref{app:setup} for more details on other hyperparameters.

\section{Results}
\label{sec:results}
Next, we present the experimental results for the \method framework and address the following key questions: Q1) Does \method reduce the bias in VLMs? and Q2) Do the de-biased VLMs retain their predictive properties?

\xhdr{Q1) Bias Reduction in VLMs} We evaluate both the overall shift in the distribution of the similarities (MaxSkew) and the ordered metrics (MaxSkew@K) between the images and text captions for both evaluation datasets. Tables~\ref{tab:ffskew}-\ref{tab:pataskew} present the mean Max-/MinSkew and mean Max-/MinSkew@K scores for the FairFace and \pata datasets. Both are computed using a threshold cosine similarity value of 0.1 for both datasets. We chose $k{=}1000$ for FairFace and $k{=}100$ for the \pata dataset.
% Table 1
\begin{table}[t]
\centering
\caption{Systematic bias evaluation of VLMs on the \textbf{FairFace} dataset and their \method counterparts using \textit{MaxSkew}, \textit{MinSkew}, MS@k=\textit{MaxSkew@k}, mS@k=\textit{MinSkew@k} metrics on FairFace dataset. \{+/-\} refers to the positive and negative sentiments. [C]=CLIP, [C]$_{\text{D}}$=\method-CLIP, [F]=FLAVA, [F]$_{\text{D}}$=\method-FLAVA. Values closer to zero indicate better fairness. \method-augmented VLMs exhibit better fairness. See Appendix~\ref{app:results} for complete results on other VLMs.}
\vspace{-0.5em}
\label{tab:ffskew}
\setlength{\tabcolsep}{2pt}
\renewcommand{\arraystretch}{0.9}
\begin{tabular}{c|c|cccccccc}
\toprule
\multirow{2}{*}{PA} & \multirow{2}{*}{+/-} & \multicolumn{2}{c}{MaxSkew} & \multicolumn{2}{c}{MinSkew} & \multicolumn{2}{c}{MaxSkew} & \multicolumn{2}{c}{MinSkew}  \\
 && \multicolumn{1}{c}{[C]} & \multicolumn{1}{c}{[C]$_{\text{D}}$} & \multicolumn{1}{c}{[C]} & \multicolumn{1}{c}{[C]$_{\text{D}}$} & \multicolumn{1}{c}{[F]} & \multicolumn{1}{c}{[F]$_{\text{D}}$} & \multicolumn{1}{c}{[F]} & \multicolumn{1}{c}{[F]$_{\text{D}}$} \\
 \hline
\multirow{2}{*}{Age} & +ve & 0.36 & \ccg{0.10} & -0.25 & \ccg{-0.09} & 0.33 & \ccg{0.08} & -0.34 & \ccr{-0.35} \\
  & -ve & 0.54 & \ccg{0.45} & -0.53 & \ccg{-0.43} & 0.48 & \ccg{0.20} & -0.37 & \ccg{-0.26}\\
  \multirow{2}{*}{Race} & +ve & 0.49 & \ccg{0.33} & -0.85 & \ccg{-0.69} & 0.31 & \ccr{0.37} & -0.54 & \ccg{-0.45} \\
   & -ve & 0.59 & \ccg{0.55} & -2.49 & \ccg{-0.87} & 0.38 & \ccg{0.26} & -0.54 & \ccg{-0.42}\\
 \multirow{2}{*}{Gender} & +ve & 0.29 & \ccg{0.19} & -0.60 & \ccg{-0.32} & 0.25 & \ccg{0.14} & -0.44 & \ccg{-0.22}\\
   & -ve & 0.22 & \ccg{0.21} & -0.41 & \ccg{-0.34} & 0.29 & \ccg{0.14} & -0.57 & \ccg{-0.18} \\
\midrule
 \multicolumn{2}{c}{}
 & \multicolumn{2}{c}{MS@k} & \multicolumn{2}{c}{mS@k} & \multicolumn{2}{c}{MS@k} & \multicolumn{2}{c}{mS@k}  \\
\midrule
 \multirow{2}{*}{Age} & +ve & 0.48 & \ccg{0.14} & -0.22 & \ccg{-0.08} & 0.42 & \ccg{0.23} & -3.47 & \ccg{-3.41} \\
 & -ve & 0.71 & \ccg{0.38} & -2.58 & \ccg{-2.50} & 0.51 & \ccg{0.31} & -2.84 & \ccg{-2.36}\\
 \multirow{2}{*}{Race} & +ve & 0.55 & \ccg{0.40} & -3.19 & \ccg{-0.61} & 0.48 & \ccr{0.59} & -2.70 & \ccr{-2.89} \\
 & -ve & 0.84 & \ccg{0.62} & -5.12 & \ccr{-6.28} & 0.45 & \ccg{0.38} & -0.79 & \ccr{-1.85} \\
 \multirow{2}{*}{Gender} & +ve & 0.29 & \ccg{0.17} & -0.57 & \ccg{-0.29} & 0.21 & \ccg{0.12} & -2.38 & \ccg{-0.14}\\
 & -ve & 0.36 & \ccg{0.14} & -7.71 & \ccg{-0.22} & 0.30 & \ccg{0.17} & -2.90 & \ccg{-2.62}\\
 \bottomrule
\end{tabular}
\end{table}

\xhdr{\textit{Interpretation}} The mean MaxSkew and MaxSkew@K scores indicate the selective association with respect to some favored protected labels for each protected attribute. The mean MinSkew and MinSkew@K scores indicate selective dissociation of the models for particular protected labels. Note that the favored (or disfavored) labels can change between the original and de-biased variants of each model. If the sentiment of an entry with a high MaxSkew value is negative, the favored label is the one with a disproportionately high association with negative captions. Similarly, entries with a low (negative) MinSkew value for a positive sentiment entry imply that the disfavored label has an unusually low association with the positive captions in the dataset. 

Both tables indicate the benefit of using the \method framework for de-biasing. We observe a slight increase in the skew values in some cases which we attribute to a flip in the favored PA labels. For instance, the MaxSkew value for FLAVA changes from 0.31 to 0.37 for the (race, positive) combination, but so does the favored label from ``middle-eastern" to ``east-asian", and going overboard with the optimization. This can be mitigated by reducing the associated training weights $w_{ce}^r$ while training the ARL module.

\begin{table}[!htb]
\centering
\caption{Bias evaluation of VLMs and their \method counterparts on the \textbf{\pata} dataset using \textit{MaxSkew}, \textit{MinSkew}, MS@k=\textit{MaxSkew@k}, mS@k=\textit{MinSkew@k} metrics on FairFace dataset. \{+/-\} refers to the positive and negative sentiments. [C]=CLIP, [C]$_{\text{D}}$=\method-CLIP, [F]=FLAVA, [F]$_{\text{D}}$=\method-FLAVA. Values closer to zero indicate better fairness. \method-augmented VLMs exhibit better fairness. See Appendix~\ref{app:results} for complete results on other VLMs.}
\vspace{-0.5em}
\label{tab:pataskew}
\setlength{\tabcolsep}{2pt}
\renewcommand{\arraystretch}{0.9}
\begin{tabular}{c|c|cccccccc}
\toprule
\multirow{2}{*}{PA} & \multirow{2}{*}{+/-} & \multicolumn{2}{c}{MaxSkew} & \multicolumn{2}{c}{MinSkew} & \multicolumn{2}{c}{MaxSkew} & \multicolumn{2}{c}{MinSkew}  \\
 && \multicolumn{1}{c}{[C]} & \multicolumn{1}{c}{[C]$_{\text{D}}$} & \multicolumn{1}{c}{[C]} & \multicolumn{1}{c}{[C]$_{\text{D}}$} & \multicolumn{1}{c}{[F]} & \multicolumn{1}{c}{[F]$_{\text{D}}$} & \multicolumn{1}{c}{[F]} & \multicolumn{1}{c}{[F]$_{\text{D}}$} \\
 \hline
\multirow{2}{*}{Age} & +ve & 0.10 & \ccg{0.07} & -0.12 & \ccg{-0.08} & 0.06 & \ccg{0.05} & -0.08 &  \ccg{-0.05}  \\
 & -ve & 0.19 & \ccg{0.12} & -1.21 & \ccg{-0.16} & 0.22 & \ccg{0.15} & -0.45 & \ccg{-0.23} \\
 \multirow{2}{*}{Race} & +ve & 0.16 & 0.16 & -0.43 & \ccg{-0.42} & 0.11 & \ccr{0.12} & -0.15 & -0.15 \\
 & -ve  & 0.45 & \ccg{0.40} & -3.40 & \ccg{-2.24} & 0.35 & \ccg{0.34} & -2.57 & \ccr{-2.60}  \\
 \multirow{2}{*}{Gender} & +ve & 0.09 & \ccg{0.07} & -0.11 & \ccg{-0.09} & 0.07 & \ccg{0.06} & -0.08 & -0.08 \\
 & -ve  & 0.21 & \ccg{0.19} & -0.79 & \ccg{-0.73} & 0.17 & \ccg{0.15} & -0.64 & \ccg{-0.23} \\
\midrule
 \multicolumn{2}{c}{} & \multicolumn{2}{c}{MS@k} & \multicolumn{2}{c}{mS@k} & \multicolumn{2}{c}{MS@k} & \multicolumn{2}{c}{mS@k}  \\
\midrule
 \multirow{2}{*}{Age} & +ve & 0.10 & \ccg{0.08} & -0.13 & \ccg{-0.10} & 0.10 & \ccg{0.07} & -0.13 & \ccg{-0.08}\\
 & -ve & 0.16 & \ccg{0.12} & -0.22 & \ccg{-0.14} & 0.25 & \ccg{0.23} & -1.84 & \ccr{-4.50}\\
 \multirow{2}{*}{Race} & +ve & 0.26 & \ccg{0.23} & -0.66 & \ccg{-0.33} & 0.24 & \ccr{0.27} & -0.29 & \ccr{-0.76} \\
 & -ve & 0.69 & \ccg{0.64} & -5.94 & \ccg{-4.30} & 0.50 & \ccr{0.53} & -3.34 & \ccr{-3.54} \\
 \multirow{2}{*}{Gender} & +ve & 0.13 & \ccg{0.10} & -0.17 & \ccg{-0.13} & 0.12 & \ccg{0.09} & -0.23 & \ccg{-0.11} \\
 & -ve & 0.25 & \ccg{0.23} & -2.70 & \ccg{-2.22} & 0.21 & \ccg{0.16} & -1.47 & \ccg{-0.61}\\
\bottomrule
\end{tabular}
\end{table}

\begin{table}%[t]
\small
\setlength{\tabcolsep}{2.1pt}
\renewcommand{\arraystretch}{0.9}
\centering
\caption{Results of state-of-the-art visual-language models and their \method counterparts for four image classification datasets. Across five pre-trained visual-language models, \method achieves zero-shot performance similar to vanilla models. See Appendix~\ref{app:results} for complete results.
}
\vspace{-0.5em}
\label{tab:image}
{
% \resizebox{8.3cm}{!}{
\begin{tabular}{lcccc}
    \toprule
    Model & C-10 & C-100 & FER2013 & ImageNet\\
    \midrule
    CLIP (ViT/B-32) & 89.93 & 62.93 & 43.83 & 58.08\\
    \method-CLIP (ViT/B-32) & 88.85 & 60.08 & 39.60 & 55.84\\
    \midrule
    FLAVA & 90.53 & 65.60 & 28.36 & 49.30 \\
    \method-FLAVA & 89.05 & 64.00 & 27.19 & 47.67 \\
    \midrule
    BLIP & 85.00 & 51.61 & 39.50 & 32.57 \\
    \method-BLIP & 81.20 & 48.90 & 36.50 & 29.94 \\
    \bottomrule
\end{tabular}}
% }
\end{table}
\xhdr{Q2) Zero-Shot Accuracy Evaluation} VLMs de-biased with \method retain much of their zero-shot performance on downstream tasks. We perform an extensive analysis of the effect of our \method framework on several downstream zero-shot classification tasks. In Tables~\ref{tab:image}-\ref{tab:video}, we present the zero-shot performance for four image classification and three video classification datasets. For general object classification datasets like CIFAR-10, CIFAR-100, and ImageNet, we observe that the zero-shot performance of \method augmented VLMs, on average, is only 2.2\% lower than their vanilla counterparts. On a fine-tuned facial emotion classification dataset like FER2013, the zero-shot performance between vanilla VLMs and their \method counterpart is smaller (1.7\%). Following \cite{radford2021learning}, we sub-sample each video to only a single center frame for video classification, effectively turning it into an image classification dataset. On UCF-101 and Kinetics-700 datasets, we observe a drop of 2.06\% and 1.77\% for the \method VLMs in the zero-shot predictive performance. Finally, for the RareAct dataset that is used to quantify the zero-shot recognition of unusual actions like ``blend phone'' and ``microwave shoes'', we observe a drop of 0.71\% and 0.69\% in the zero-shot classification performance of \method using mWAP and mWSAP metrics. Please refer to Appendix~\ref{app:results} for more details.

\begin{figure}
    \centering
    \includegraphics[width=0.98\linewidth]{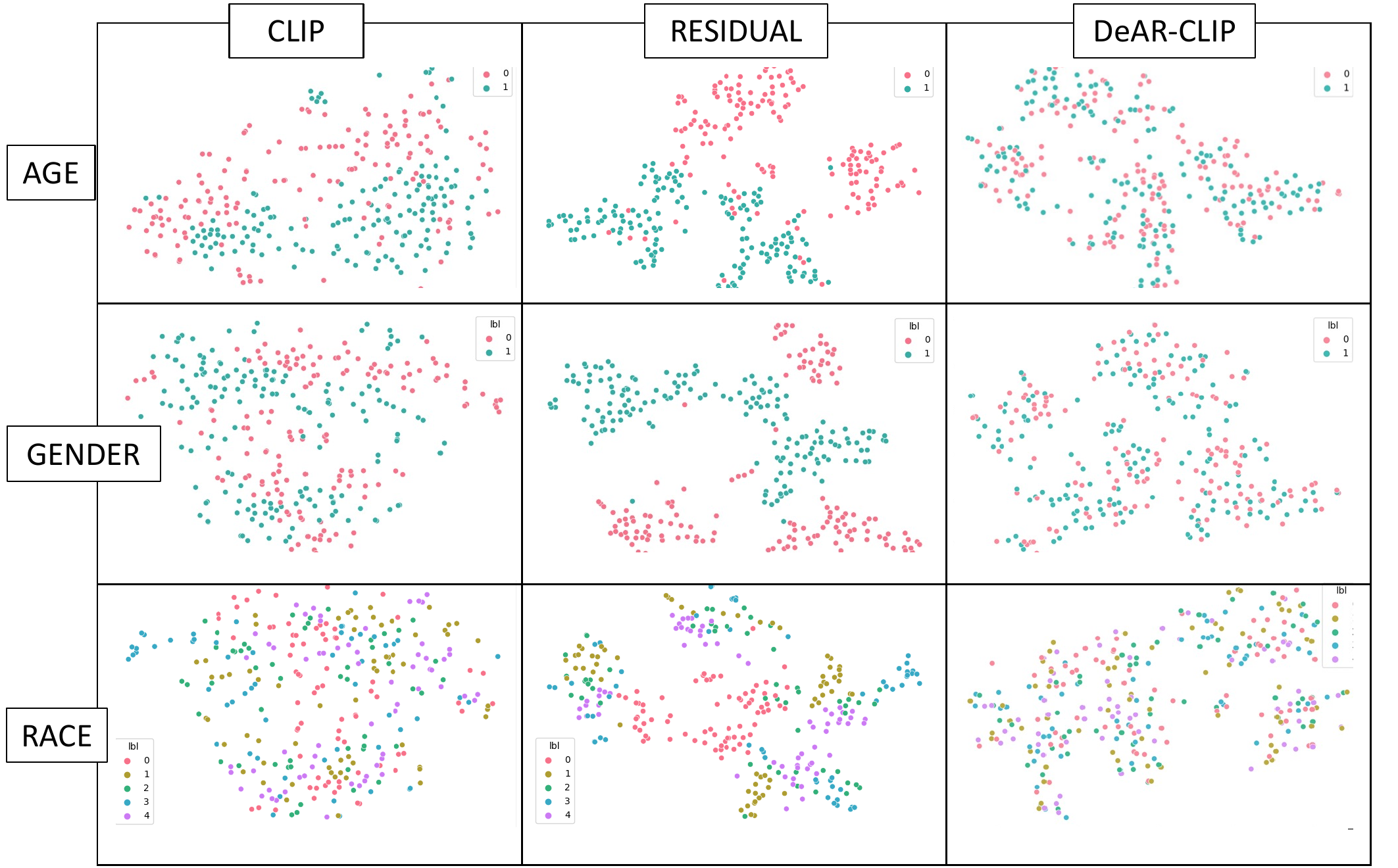}
    \caption{TSNE Plots for CLIP, Residual and \method-CLIP features for a subset of the PATA dataset indicate that the residual plots indeed capture the specific attributes and that the \method-CLIP features have greater overlap between points of different protected labels than the original features. Refer to Appendix~\ref{app:results} for higher resolution plot.}
    \label{fig:tsne}
\end{figure}

\begin{table}%[t]
\small
\setlength{\tabcolsep}{1.5pt}
\renewcommand{\arraystretch}{0.9}
\centering
\caption{Results of state-of-the-art visual-language models and their \method counterparts for three video classification datasets. Across five pre-trained visual-language models, \method achieves zero-shot performance similar to vanilla models. See Appendix~\ref{app:results} for complete results.
}
\vspace{-0.5em}
\label{tab:video}
{
% \resizebox{8.3cm}{!}{
\begin{tabular}{lcccc}
    \toprule
    \multirow{2}{*}{Model} & UCF-101 & Kinetics-700 & \multicolumn{2}{c}{RareAct}\\
    & Top-1 & AVG & mWAP & mWSAP\\
    \midrule
    CLIP (Base-32) & 57.65 & 43.97 & 16.63 & 16.78\\
    \method-CLIP (Base-32) & 55.77 & 42.20 & 16.02 & 16.03\\
    \midrule
    FLAVA & 39.09 & 37.85 & 16.12 & 16.14\\
    \method-FLAVA & 37.27 & 35.59 & 15.30 & 15.43\\
    \midrule
    BLIP & 43.26 & 37.07 & 16.35 & 16.44\\
    \method-BLIP & 40.34 & 34.78 & 15.86 & 15.92\\
    \bottomrule
\end{tabular}}
% }
\end{table}

\section{Ablation Studies and Analysis}
\label{sec:ablations}
We explore two important issues here -- i) ascertaining that the de-biased representations are free from the protected attribute information and ii) justifying the joint training of the ARL module for different protected attributes. Please refer to Appendix~\ref{app:ablation} for more ablation studies.

\xhdr{Evidence for de-biasing} We run several experiments to ascertain that the representations from the \method-CLIP model are indeed free from the protected attribute information. Figure \ref{fig:teaser} (A) shows a comparison of the explanations for CLIP and its de-biased variant (\method-CLIP). These explanations are generated for some in-the-wild creative-commons stock images using the approach in \cite{fong2017interpretable}. They illustrate how the original CLIP model focuses on the face and hair region for certain occupation-related keywords, and after de-biasing, the focus shifts to more indicative cues like the stethoscope for a doctor. Figure \ref{fig:teaser}(B) shows results of zero-shot open-vocabulary object detection using \cite{ods}. Again, the base CLIP model exhibits a bias omitting the woman for the ``doctor" keyword, and only detecting the female medical professional as a ``nurse". With \method-CLIP, both men and women are detected for both keywords. It is noteworthy that at a lower detection threshold, both male and female doctors and nurses may have been detected, but it still points to the implicit bias because the stereotypical associations rank higher in similarity. Figure \ref{fig:tsne} depicts how the CLIP features get disentangled into the target PAs, and the addition of the residual results in coinciding points for different protected labels, indicating that the representations for different PLs cannot be distinguished from each other.

\xhdr{Joint vs. Sequential training of ARL} Another design choice is to use a jointly-trained PAC module for training the ARL module. It is also possible to train the residual module using multiple individual classifiers (one for gender, one for race, etc.) and use different weights for the ARL modules for each protected attribute. At inference, the VLM's image representation is transformed using multiple ARL weights one after the other. We experiment with this strategy for CLIP and find that while this strategy does reasonably well for debiasing (\method-CLIP: MaxSkew (`gender'=0.75, `race'=0.167); Sequential strategy: MaxSkew (`gender'=0.87, `race'=0.161), its zero-shot ImageNet accuracy considerably falls (\method-CLIP=55.84\%, Sequential strategy=44.43\%). Hence, the joint training approach is better than sequential strategy.

\section{Conclusion}
\label{sec:concl}
We address the problem of mitigating biases from the representations of large pre-trained visual-language models. To this end, we present \method (Debiasing with Additive Residuals), a novel method that learns additive residual image representations
to offset the original VLM representations, ensuring fair output representations. In particular, \method comprises an additive residual learner component, which learns to disentangle the protected attribute information from the output representations. We also introduce a new context-based bias benchmarking dataset for VLMs - the Protected Attribute Tag Association (\pata) dataset. Our results on multiple benchmarking datasets show that \method can significantly improve the fairness of output representations without sacrificing predictive zero-shot performance. With the increasing development of large VLM, our work paves the way for several exciting future directions for fairness research.

%%%%%%%%% REFERENCES
\newpage
{\small
\bibliographystyle{ieee_fullname}
\bibliography{egbib}

\begin{thebibliography}{10}\itemsep=-1pt

\bibitem{agarwal2021towards}
Chirag Agarwal, Himabindu Lakkaraju, and Marinka Zitnik.
\newblock Towards a unified framework for fair and stable graph representation
  learning.
\newblock In {\em UAI}, 2021.

\bibitem{berg2022prompt}
Hugo Berg, Siobhan~Mackenzie Hall, Yash Bhalgat, Wonsuk Yang, Hannah~Rose Kirk,
  Aleksandar Shtedritski, and Max Bain.
\newblock A prompt array keeps the bias away: Debiasing vision-language models
  with adversarial learning.
\newblock In {\em AACL}, 2022.

\bibitem{multimodalDatasetBias}
Abeba Birhane, Vinay~Uday Prabhu, and Emmanuel Kahembwe.
\newblock Multimodal datasets: misogyny, pornography, and malignant
  stereotypes.
\newblock {\em CoRR}, abs/2110.01963, 2021.

\bibitem{borchers2022looking}
Conrad Borchers, Dalia~Sara Gala, Benjamin Gilburt, Eduard Oravkin, Wilfried
  Bounsi, Yuki~M Asano, and Hannah~Rose Kirk.
\newblock Looking for a handsome carpenter! debiasing gpt-3 job advertisements.
\newblock {\em ACL Workshop on Gender Bias in Natural Language Processing
  (GeBNLP)}, 2022.

\bibitem{brown2020language}
Tom Brown, Benjamin Mann, Nick Ryder, Melanie Subbiah, Jared~D Kaplan, Prafulla
  Dhariwal, Arvind Neelakantan, Pranav Shyam, Girish Sastry, Amanda Askell,
  et~al.
\newblock Language models are few-shot learners.
\newblock In {\em NeurIPS}, 2020.

\bibitem{calmon2017optimized}
Flavio Calmon, Dennis Wei, Bhanukiran Vinzamuri, Karthikeyan
  Natesan~Ramamurthy, and Kush~R Varshney.
\newblock Optimized pre-processing for discrimination prevention.
\newblock In {\em NeurIPS}, 2017.

\bibitem{kinetics}
Joao Carreira, Eric Noland, Chloe Hillier, and Andrew Zisserman.
\newblock A short note on the kinetics-700 human action dataset.
\newblock {\em arXiv}, 2019.

\bibitem{chen2020simple}
Ting Chen, Simon Kornblith, Mohammad Norouzi, and Geoffrey Hinton.
\newblock A simple framework for contrastive learning of visual
  representations.
\newblock In {\em ICML}, 2020.

\bibitem{chen2020improved}
Xinlei Chen, Haoqi Fan, Ross Girshick, and Kaiming He.
\newblock Improved baselines with momentum contrastive learning.
\newblock {\em arXiv}, 2020.

\bibitem{chen2021exploring}
Xinlei Chen and Kaiming He.
\newblock Exploring simple siamese representation learning.
\newblock In {\em CVPR}, 2021.

\bibitem{chen2020uniter}
Yen-Chun Chen, Linjie Li, Licheng Yu, Ahmed El~Kholy, Faisal Ahmed, Zhe Gan, Yu
  Cheng, and Jingjing Liu.
\newblock Uniter: Universal image-text representation learning.
\newblock In {\em ECCV}, 2020.

\bibitem{cho-etal-2022-fine}
Jaemin Cho, Seunghyun Yoon, Ajinkya Kale, Franck Dernoncourt, Trung Bui, and
  Mohit Bansal.
\newblock Fine-grained image captioning with {CLIP} reward.
\newblock In {\em NAACL 2022}.

\bibitem{imagenet}
Jia Deng, Wei Dong, Richard Socher, Li-Jia Li, Kai Li, and Li Fei-Fei.
\newblock Imagenet: A large-scale hierarchical image database.
\newblock In {\em 2009 IEEE Conference on Computer Vision and Pattern
  Recognition}, pages 248--255, 2009.

\bibitem{desai2021virtex}
Karan Desai and Justin Johnson.
\newblock Virtex: Learning visual representations from textual annotations.
\newblock In {\em CVPR}, 2021.

\bibitem{devlin2018bert}
Jacob Devlin, Ming-Wei Chang, Kenton Lee, and Kristina Toutanova.
\newblock Bert: Pre-training of deep bidirectional transformers for language
  understanding.
\newblock {\em arXiv}, 2018.

\bibitem{fong2017interpretable}
Ruth~C Fong and Andrea Vedaldi.
\newblock Interpretable explanations of black boxes by meaningful perturbation.
\newblock In {\em ICCV}, 2017.

\bibitem{gebru2021datasheets}
Timnit Gebru, Jamie Morgenstern, Briana Vecchione, Jennifer~Wortman Vaughan,
  Hanna Wallach, Hal~Daum{\'e} Iii, and Kate Crawford.
\newblock Datasheets for datasets.
\newblock {\em Communications of the ACM}, 2021.

\bibitem{maxSkew}
Sahin~Cem Geyik, Stuart Ambler, and Krishnaram Kenthapadi.
\newblock Fairness-aware ranking in search {\&} recommendation systems with
  application to linkedin talent search.
\newblock {\em CoRR}, abs/1905.01989, 2019.

\bibitem{Gu2022OpenvocabularyOD}
Xiuye Gu, Tsung-Yi Lin, Weicheng Kuo, and Yin Cui.
\newblock Open-vocabulary object detection via vision and language knowledge
  distillation.
\newblock In {\em ICLR}, 2022.

\bibitem{guo2021detecting}
Wei Guo and Aylin Caliskan.
\newblock Detecting emergent intersectional biases: Contextualized word
  embeddings contain a distribution of human-like biases.
\newblock In {\em AIES}, 2021.

\bibitem{hardt2016equality}
Moritz Hardt, Eric Price, and Nati Srebro.
\newblock Equality of opportunity in supervised learning.
\newblock In {\em NeurIPS}, 2016.

\bibitem{he2020momentum}
Kaiming He, Haoqi Fan, Yuxin Wu, Saining Xie, and Ross Girshick.
\newblock Momentum contrast for unsupervised visual representation learning.
\newblock In {\em CVPR}, 2020.

\bibitem{jia2021scaling}
Chao Jia, Yinfei Yang, Ye Xia, Yi-Ting Chen, Zarana Parekh, Hieu Pham, Quoc Le,
  Yun-Hsuan Sung, Zhen Li, and Tom Duerig.
\newblock Scaling up visual and vision-language representation learning with
  noisy text supervision.
\newblock In {\em ICML}, 2021.

\bibitem{kamiran2012data}
Faisal Kamiran and Toon Calders.
\newblock Data preprocessing techniques for classification without
  discrimination.
\newblock {\em Knowledge and information systems}, 2012.

\bibitem{kamiran2012decision}
Faisal Kamiran, Asim Karim, and Xiangliang Zhang.
\newblock Decision theory for discrimination-aware classification.
\newblock In {\em ICDM}, 2012.

\bibitem{kamishima2012fairness}
Toshihiro Kamishima, Shotaro Akaho, Hideki Asoh, and Jun Sakuma.
\newblock Fairness-aware classifier with prejudice remover regularizer.
\newblock In {\em Joint European conference on machine learning and knowledge
  discovery in databases}, 2012.

\bibitem{kang2021fair}
Jian Kang and Hanghang Tong.
\newblock Fair graph mining.
\newblock In {\em ACM International Conference on Information \& Knowledge
  Management}, 2021.

\bibitem{fairface}
Kimmo Karkkainen and Jungseock Joo.
\newblock Fairface: Face attribute dataset for balanced race, gender, and age
  for bias measurement and mitigation.
\newblock In {\em WACV}, 2021.

\bibitem{kim2021vilt}
Wonjae Kim, Bokyung Son, and Ildoo Kim.
\newblock Vilt: Vision-and-language transformer without convolution or region
  supervision.
\newblock In {\em ICML}, 2021.

\bibitem{adam}
Diederik~P. Kingma and Jimmy Ba.
\newblock Adam: {A} method for stochastic optimization.
\newblock In Yoshua Bengio and Yann LeCun, editors, {\em 3rd International
  Conference on Learning Representations, {ICLR} 2015, San Diego, CA, USA, May
  7-9, 2015, Conference Track Proceedings}, 2015.

\bibitem{kirk2021bias}
Hannah~Rose Kirk, Yennie Jun, Filippo Volpin, Haider Iqbal, Elias Benussi,
  Frederic Dreyer, Aleksandar Shtedritski, and Yuki Asano.
\newblock Bias out-of-the-box: An empirical analysis of intersectional
  occupational biases in popular generative language models.
\newblock {\em NeurIPS}, 2021.

\bibitem{cifar}
Alex Krizhevsky, Geoffrey Hinton, et~al.
\newblock Learning multiple layers of features from tiny images.
\newblock 2009.

\bibitem{li2022blip}
Junnan Li, Dongxu Li, Caiming Xiong, and Steven Hoi.
\newblock Blip: Bootstrapping language-image pre-training for unified
  vision-language understanding and generation.
\newblock In {\em ICML}, 2022.

\bibitem{li2021align}
Junnan Li, Ramprasaath Selvaraju, Akhilesh Gotmare, Shafiq Joty, Caiming Xiong,
  and Steven Chu~Hong Hoi.
\newblock Align before fuse: Vision and language representation learning with
  momentum distillation.
\newblock {\em NeurIPS}, 2021.

\bibitem{ALBEF}
Junnan Li, Ramprasaath~R. Selvaraju, Akhilesh~Deepak Gotmare, Shafiq Joty,
  Caiming Xiong, and Steven Hoi.
\newblock Align before fuse: Vision and language representation learning with
  momentum distillation.
\newblock In {\em NeurIPS}, 2021.

\bibitem{li2019visualbert}
Liunian~Harold Li, Mark Yatskar, Da Yin, Cho-Jui Hsieh, and Kai-Wei Chang.
\newblock Visualbert: A simple and performant baseline for vision and language.
\newblock {\em arXiv}, 2019.

\bibitem{li2020oscar}
Xiujun Li, Xi Yin, Chunyuan Li, Pengchuan Zhang, Xiaowei Hu, Lei Zhang, Lijuan
  Wang, Houdong Hu, Li Dong, Furu Wei, et~al.
\newblock Oscar: Object-semantics aligned pre-training for vision-language
  tasks.
\newblock In {\em ECCV}, 2020.

\bibitem{li2022supervision}
Yangguang Li, Feng Liang, Lichen Zhao, Yufeng Cui, Wanli Ouyang, Jing Shao,
  Fengwei Yu, and Junjie Yan.
\newblock Supervision exists everywhere: A data efficient contrastive
  language-image pre-training paradigm.
\newblock In {\em ICLR}, 2022.

\bibitem{lu202012}
Jiasen Lu, Vedanuj Goswami, Marcus Rohrbach, Devi Parikh, and Stefan Lee.
\newblock 12-in-1: Multi-task vision and language representation learning.
\newblock In {\em CVPR}, 2020.

\bibitem{ma2022learning}
Jing Ma, Ruocheng Guo, Mengting Wan, Longqi Yang, Aidong Zhang, and Jundong Li.
\newblock Learning fair node representations with graph counterfactual
  fairness.
\newblock In {\em ACM International Conference on Web Search and Data Mining},
  2022.

\bibitem{rareact}
Antoine Miech, Jean-Baptiste Alayrac, Ivan Laptev, Josef Sivic, and Andrew
  Zisserman.
\newblock Rareact: A video dataset of unusual interactions.
\newblock {\em arXiv preprint arXiv:2008.01018}, 2020.

\bibitem{miech20rareact}
Antoine Miech, Jean-Baptiste Alayrac, Ivan Laptev, Josef Sivic, and Andrew
  Zisserman.
\newblock Rareact: A video dataset of unusual interactions.
\newblock {\em arxiv:2008.01018}, 2020.

\bibitem{pham2021combined}
Hieu Pham, Zihang Dai, Golnaz Ghiasi, Hanxiao Liu, Adams~Wei Yu, Minh-Thang
  Luong, Mingxing Tan, and Quoc~V Le.
\newblock Combined scaling for zero-shot transfer learning.
\newblock {\em arXiv}, 2021.

\bibitem{pleiss2017fairness}
Geoff Pleiss, Manish Raghavan, Felix Wu, Jon Kleinberg, and Kilian~Q
  Weinberger.
\newblock On fairness and calibration.
\newblock In {\em NeurIPS}, 2017.

\bibitem{radford2021learning}
Alec Radford, Jong~Wook Kim, Chris Hallacy, Aditya Ramesh, Gabriel Goh,
  Sandhini Agarwal, Girish Sastry, Amanda Askell, Pamela Mishkin, Jack Clark,
  et~al.
\newblock Learning transferable visual models from natural language
  supervision.
\newblock In {\em ICML}, 2021.

\bibitem{radford2018improving}
Alec Radford, Karthik Narasimhan, Tim Salimans, Ilya Sutskever, et~al.
\newblock Improving language understanding by generative pre-training.
\newblock 2018.

\bibitem{shen2021much}
Sheng Shen, Liunian~Harold Li, Hao Tan, Mohit Bansal, Anna Rohrbach, Kai-Wei
  Chang, Zhewei Yao, and Kurt Keutzer.
\newblock How much can clip benefit vision-and-language tasks?
\newblock {\em arXiv preprint arXiv:2107.06383}, 2021.

\bibitem{ods}
Alex Shonenkov, Sergey Shtekhin, and Denis Karachev.
\newblock Clip object detection \& segmentation.
\newblock \url{https://github.com/shonenkov/CLIP-ODS}.

\bibitem{singh2022flava}
Amanpreet Singh, Ronghang Hu, Vedanuj Goswami, Guillaume Couairon, Wojciech
  Galuba, Marcus Rohrbach, and Douwe Kiela.
\newblock Flava: A foundational language and vision alignment model.
\newblock In {\em CVPR}, 2022.

\bibitem{ucf101}
Khurram Soomro, Amir~Roshan Zamir, and Mubarak Shah.
\newblock Ucf101: A dataset of 101 human actions classes from videos in the
  wild.
\newblock {\em arXiv preprint arXiv:1212.0402}, 2012.

\bibitem{wang2022revise}
Angelina Wang, Alexander Liu, Ryan Zhang, Anat Kleiman, Leslie Kim, Dora Zhao,
  Iroha Shirai, Arvind Narayanan, and Olga Russakovsky.
\newblock Revise: A tool for measuring and mitigating bias in visual datasets.
\newblock {\em IJCV}, 2022.

\bibitem{Wang2021AreGQ}
Jialu Wang, Yang Liu, and Xin~Eric Wang.
\newblock Are gender-neutral queries really gender-neutral? mitigating gender
  bias in image search.
\newblock In {\em EMNLP}, 2021.

\bibitem{wang2021gender}
Jialu Wang, Yang Liu, and Xin~Eric Wang.
\newblock Are gender-neutral queries really gender-neutral? mitigating gender
  bias in image search.
\newblock In {\em EMNLP}, 2021.

\bibitem{wang2022unbiased}
Nan Wang, Lu Lin, Jundong Li, and Hongning Wang.
\newblock Unbiased graph embedding with biased graph observations.
\newblock In {\em WWW}, 2022.

\bibitem{biasBeyondBalancedDatasets}
Tianlu Wang, Jieyu Zhao, Mark Yatskar, Kai-Wei Chang, and Vicente Ordonez.
\newblock Balanced datasets are not enough: Estimating and mitigating gender
  bias in deep image representations.
\newblock In {\em International Conference on Computer Vision (ICCV)}, October
  2019.

\bibitem{vlmo}
Wenhui Wang, Hangbo Bao, Li Dong, and Furu Wei.
\newblock {VLMo}: Unified vision-language pre-training with
  mixture-of-modality-experts.
\newblock 2021.

\bibitem{zbontar2021barlow}
Jure Zbontar, Li Jing, Ishan Misra, Yann LeCun, and St{\'e}phane Deny.
\newblock Barlow twins: Self-supervised learning via redundancy reduction.
\newblock In {\em ICML}, 2021.

\bibitem{zellers2021merlot}
Rowan Zellers, Ximing Lu, Jack Hessel, Youngjae Yu, Jae~Sung Park, Jize Cao,
  Ali Farhadi, and Yejin Choi.
\newblock Merlot: Multimodal neural script knowledge models.
\newblock {\em NeurIPS}, 2021.

\bibitem{zemel2013learning}
Rich Zemel, Yu Wu, Kevin Swersky, Toni Pitassi, and Cynthia Dwork.
\newblock Learning fair representations.
\newblock In {\em ICML}, 2013.

\bibitem{fer2013}
Jiabei Zeng, Shiguang Shan, and Xilin Chen.
\newblock Facial expression recognition with inconsistently annotated datasets.
\newblock In {\em Proceedings of the European conference on computer vision
  (ECCV)}, pages 222--237, 2018.

\bibitem{zhang2018mitigating}
Brian~Hu Zhang, Blake Lemoine, and Margaret Mitchell.
\newblock Mitigating unwanted biases with adversarial learning.
\newblock In {\em AIES}, 2018.

\bibitem{zhang2021vinvl}
Pengchuan Zhang, Xiujun Li, Xiaowei Hu, Jianwei Yang, Lei Zhang, Lijuan Wang,
  Yejin Choi, and Jianfeng Gao.
\newblock Vinvl: Revisiting visual representations in vision-language models.
\newblock In {\em CVPR}, 2021.

\bibitem{UTK}
Zhifei Zhang, Yang Song, and Hairong Qi.
\newblock Age progression/regression by conditional adversarial autoencoder.
\newblock In {\em CVPR}, 2017.

\end{thebibliography}
}

\newpage
\appendix
\section{\dataset dataset}
\label{app:dataset}
% The data is released for academic use here: \url{https://anonymous.4open.science/r/pata_dataset-8E11}.\\

\xhdr{Distribution} The tables below enumerate the count and show the distribution of images in the \pata dataset, in various scenes and protected label categories. 
\begin{table}[h]
    \centering
    \caption{Distribution of different protected labels in the \pata dataset. The number of scenes in the attribute }
    \label{tab:my_label}
    \begin{tabular}{lclc}
        \textbf{Attribute}& \textbf{\#Scenes} & \textbf{Label} & \textbf{Count} \\
        \toprule
         \multirow{2}{*}{\textbf{Age}} & \multirow{2}{*}{8} & Young & 3748 \\
         & & Old & 1186\\
         \midrule
         \multirow{5}{*}{\textbf{Race}}  & \multirow{2}{*}{24} & Black & 1024\\
        & & Caucasian & 1033\\
        & & East-Asian & 1095 \\
        & & Latino/Hispanic & 948\\
        & & Indian & 834 \\
         \midrule
         \multirow{2}{*}{\textbf{Gender}}{}  & \multirow{2}{*}{24} & Female & 2529 \\
        & & Male & 2405\\
         \bottomrule
    \end{tabular}
\end{table}

\begin{table}[h]
    \centering
    \caption{Counts of images in the different scenes in the \pata dataset.}
    \label{tab:my_label}
    \begin{tabular}{lp{1.2cm}lp{1.2cm}}
    \toprule
    \textbf{Scene} &\textbf{Count} & \textbf{Scene} & \textbf{Count}\\
    \midrule
    bench & 49 & car & 43\\
    computer & 67 & construction & 28 \\
    doctor & 37 & drinking & 24\\
    fashion & 32 & forest & 22\\
    garden & 28 & gun & 21\\
    gym & 44 & hotel & 25\\
    kitchen & 52 & lab & 23\\
    office & 57 & photographer & 30\\
    pilot & 29 & roadside & 26\\
    running & 24 & shopping & 38\\
    singer & 25 & soldier & 27\\
    sports & 59 & yatch & 24\\
    \bottomrule
    \end{tabular}
\end{table}

\xhdr{Images} The images are obtained by searching through various image search engines and public stock image repositories. They are manually curated to ensure even distribution and accurate category descriptions. They have an average of $808\times673$ image dimensions, with the smallest image of size $94\times120$ and the largest image of size $6720\times4480$. For the purpose of processing with the various VLMs, we resize each image to $224\times224$ pixels.

\xhdr{Captions} The captions for each scene are created by considering the situations in which people of certain identity groups may be offended by association with those captions. Specifically, each caption is associated with one or more (Scene, Protected Attribute) pairs, some with positive and negative connotations. Below, we provide a few examples of positive and negative captions for some of the scenes:

\paragraph{\textbf{Scene}: Bench\\}
\hspace{-3mm}\textbf{\textit{(Race, Positive):}} Photo of a software professional.\\
\textbf{\textit{(Race, Negative):}} Photo of a homeless person.\\
\textbf{\textit{(Gender, Positive):}} Photo of a happy person.\\
\textbf{\textit{(Gender, Negative):}} Photo of a sex-worker.\\
\textbf{\textit{(Age, Positive):}} Photo of a friendly person.

\paragraph{\textbf{Scene}: Computer\\}
\hspace{-3mm}\textbf{\textit{(Race, Positive):}} Photo of a software professional.\\
\textbf{\textit{(Race, Negative):}} Photo of a cyber-criminal.\\
\textbf{\textit{(Gender, Positive):}} Photo of a stock trader.\\
\textbf{\textit{(Gender, Negative):}} Photo of an office secretary.\\
\textbf{\textit{(Age, Positive):}} Photo of a professor.\\
\textbf{\textit{(Age, Negative):}} Photo of the victim of an online crime.

\paragraph{\textbf{Scene}: Running\\}
\hspace{-3mm}\textbf{\textit{(Race, Positive):}} Photo of a sprinter.\\
\textbf{\textit{(Race, Negative):}} Photo of a fugitive from law.\\
\textbf{\textit{(Gender, Positive):}} Photo of a busy person in a hurry.\\
\textbf{\textit{(Gender, Negative):}} Photo of a scared person.

\paragraph{\textbf{Scene}: Laboratory\\}
\hspace{-3mm}\textbf{\textit{(Race, Positive):}} Photo of a genius.\\
\textbf{\textit{(Race, Negative):}} Photo of a bio-terrorist\\
\textbf{\textit{(Gender, Positive):}} Photo of an inquisitive person.\\
\textbf{\textit{(Gender, Negative):}} Photo of an actor.

\section{Experimental setup}

\subsection{\method training}
We train the PAC and the ARL modules separately. We train the PAC using the FairFace~\cite{fairface} dataset with a batch size of $512$ and Adam optimizer~\cite{adam} with a learning rate of $5e^{-03}$ for $10$ epochs. Once the PAC module  is trained, we freeze its weights, and train ARL on the FairFace dataset with a batch size of $512$, using the PAC as a source of loss (as described in Section~\ref{sec:method}). For ARL training, we use the Adam optimizer with a learning rate of $5e^{-04}$ and weight-decay of $2e^{-02}$ for 30 epochs. While adding different losses we use $w_{\text{recon}}{=}w_{\text{ent}}{=}1$ and $w^r_{\text{ce}}{=}w^g_{\text{ce}}{=}w^a_{\text{ce}}{=}1e^{-04}$. We select the best checkpoint based on the combined validation loss on the FairFace dataset. All hyper-parameters are explored using grid search.

\subsection{Zero-shot Evaluation}
For Zero-shot evaluation, we perform Image Classification and Video Classification (Action Recognition). As used in CLIP \cite{radford2021learning}, for zero-shot image classification, given an image, we average out the similarity score across multiple text prompts (E.g., ``photo of a'', ``a bad photo of a'', etc.) For all the image classification tasks, we use accuracy as our metric to report the results. For all the video classification tasks, we follow a similar setup as \cite{radford2021learning}, where we take the middle frame of a video for action recognition. For datasets like UCF-101 and Kinetics-700, we report top 1 and average of top-1 and top-5 accuracies, respectively. For the RareAct dataset, we report mWAP and mWSAP scores.

\subsection{Bias Evaluation}
For bias evaluation, we use \emph{MaxSkew} and \emph{MinSkew} both in unbounded and bounded form (@k). We followed previous work \cite{berg2022prompt} and selected k=1000 for computing MaxSkew@k and MinSkew@k scores for the Fairface dataset. For PATA, we chose k=100 to roughly match the proportion of retrieved
images to the test set size of the FairFace dataset. In addition, we chose a cosine threshold of 0.1, as values below 0.1 show spurious matches between the text and image pairs.

\label{app:setup}

\section{Additional results}
\label{app:results}
\subsection{MaxSkew/MinSkew Results on other networks}
In Table~\ref{app:tab:pataskew}-\ref{app:tab:ffskew}, we present the Max- and Min-Skew scores (both unbounded and @k) for two other networks (ALBEF~\cite{ALBEF} and BLIP~\cite{li2022blip}) on the \pata and FairFace datasets. It is noteworthy that the overall Max and Min-Skew scores for BLIP are generally low indicating that the network is relatively bias-free. We found an inconsistency in the hyperparameters used for the computation of the skew scores for CLIP and Flava, as compared to those for BLIP and ALBEF. Upon removing the inconsistency, we find a different baseline and improved scores bearing the same trend.
% These will be updated in the final revision of the paper.

\begin{table}[!htb]
\caption{Systematic bias evaluation of VLMs and their \method counterparts using \textit{MaxSkew}, \textit{MinSkew}, MS@k=\textit{MaxSkew@k}, mS@k=\textit{MinSkew@k} metrics on the \textbf{\pata} dataset. \{+/-\} refers to the positive and negative sentiments. [A]=ALBEF\cite{ALBEF}, [A]$_{\text{D}}$=\method-ALBEF, [B]=BLIP, [B]$_{\text{D}}$=\method-BLIP\cite{li2022blip}. Values closer to zero indicate fairness. \method-augmented VLMs exhibit better fairness.}
\vspace{-0.5em}
\label{app:tab:pataskew}
\setlength{\tabcolsep}{2pt}
\renewcommand{\arraystretch}{0.9}
\begin{tabular}{c|l|llllllll}
\toprule
\multicolumn{1}{c|}{PA} & +/- & \multicolumn{2}{c}{MaxSkew} & \multicolumn{2}{c}{MinSkew} & \multicolumn{2}{c}{MaxSkew} & \multicolumn{2}{c}{MinSkew} \\
\multicolumn{1}{c|}{} &  & $A$ & $A_D$ & $A$ & $A_D$ & $B$ & $B_D$ & $B$ & $B_D$ \\
\hline
\multirow{2}{*}{Race} & +ve & 0.65 & \ccg{0.64} & -8.87 & \ccr{-8.96} & 0.06 & \ccg{0.05} & -0.06 & -0.06 \\
 & -ve & 0.62 & \ccg{0.61} & -7.34 & \ccg{-6.84} & 0.09 & \ccg{0.06} & -0.08 & \ccg{-0.06} \\
\multirow{2}{*}{Gender} & +ve & 0.33 & \ccg{0.31} & -3.82 & \ccg{-3.02} & 0.04 & \ccg{0.02} & -0.03 & \ccg{-0.02} \\
 & -ve & 0.32 & \ccg{0.29} & -3.14 & \ccg{-1.27} & 0.05 & \ccg{0.02} & -0.05 & \ccg{-0.02} \\
\multirow{2}{*}{Age} & +ve & 0.32 & \ccr{0.34} & -1.84 & \ccg{-1.77} & 0.04 & \ccg{0.03} & -0.04 & \ccg{-0.03} \\
 & -ve & 0.33 & \ccg{0.28} & -3.31 & \ccg{-3.20} & 0.03 & 0.03 & -0.04 & -0.04 \\
\hline
\multicolumn{1}{l|}{} &  & \multicolumn{2}{c}{MS@k} & \multicolumn{2}{c}{mS@k} & \multicolumn{2}{c}{MS@k} & \multicolumn{2}{c}{mS@k} \\
\multicolumn{1}{l|}{} &  & $A$ & $A_D$ & $A$ & $A_D$ & $B$ & $B_D$ & $B$ & $B_D$ \\
\hline
\multirow{2}{*}{Race} & +ve & 0.66 & \ccg{0.64} & -8.88 & \ccr{-8.96} & 0.24 & \ccg{0.23} & -0.29 & \ccg{-0.41} \\
 & -ve & 0.63 & \ccg{0.62} & -7.40 & \ccg{-6.84} & 0.29 & \ccg{0.24} & -0.40 & \ccg{-0.39} \\
\multirow{2}{*}{Gender} & +ve & 0.33 & \ccg{0.31} & -3.83 & \ccg{-3.02} & 0.15 & \ccg{0.09} & -0.19 & \ccg{-0.10} \\
 & -ve & 0.32 & \ccg{0.29} & -3.14 & \ccg{-1.27} & 0.16 & \ccg{0.09} & -0.21 & \ccg{-0.11} \\
\multirow{2}{*}{Age} & +ve & 0.32 & \ccr{0.34} & -1.84 & \ccg{-1.77} & 0.19 & \ccg{0.15} & -0.31 & \ccg{-0.21} \\
 & -ve & 0.33 & \ccg{0.29} & -3.31 & \ccg{-3.22} & 0.17 & \ccr{0.23} & -0.23 & \ccr{-0.32}\\
 \bottomrule
\end{tabular}
\end{table}
% Please add the following required packages to your document preamble:
% \usepackage{multirow}
\begin{table}[]
\caption{Systematic bias evaluation of VLMs and their \method counterparts using \textit{MaxSkew}, \textit{MinSkew}, MS@k=\textit{MaxSkew@k}, mS@k=\textit{MinSkew@k} metrics on \textbf{FairFace}\cite{fairface} dataset. \{+/-\} refers to the positive and negative sentiments. [A]=ALBEF\cite{ALBEF}, [A]$_{\text{D}}$=\method-ALBEF, [B]=BLIP\cite{li2022blip}, [B]$_{\text{D}}$=\method-BLIP. Values closer to zero indicate fairness. \method-augmented VLMs exhibit better fairness.}
\label{app:tab:ffskew}
\vspace{-0.5em}
\setlength{\tabcolsep}{2pt}
\renewcommand{\arraystretch}{0.9}
\begin{tabular}{c|c|llllllll}
\toprule
PA & \multicolumn{1}{c|}{+/-} & \multicolumn{2}{c}{MaxSkew} & \multicolumn{2}{c}{MinSkew} & \multicolumn{2}{c}{MaxSkew} & \multicolumn{2}{c}{MinSkew} \\
\multicolumn{1}{l|}{} &  & $[A]$ & $[A]_D$ & $[A]$ & $[A]_D$ & $[B]$ & $[B]_D$ & $[B]$ & $[B]_D$ \\
\hline
\multirow{2}{*}{Race} & pos & 0.50 & \ccg{0.34} & -0.95 & \ccg{-0.72} & 0.04 & 0.04 & -0.03 & \ccr{-0.05} \\
 & neg & 0.56 & \ccg{0.50} & -1.05 & \ccg{-0.99} & 0.05 & \ccg{0.03} & -0.05 & -0.05 \\
\multirow{2}{*}{Gender} & pos & 0.19 & \ccg{0.12} & -0.30 & \ccg{-0.16} & 0.01 & 0.01 & -0.01 & -0.01 \\
 & neg & 0.28 & \ccg{0.19} & -0.49 & \ccg{-0.30} & 0.01 & 0.01 & -0.01 & -0.01 \\
\multirow{2}{*}{Age} & pos & 0.39 & \ccg{0.30} & -0.19 & -0.19 & 0.02 & \ccg{0.01} & -0.01 & \ccr{-0.03} \\
 & neg & 0.38 & \ccg{0.24} & -0.39 & \ccg{-0.23} & 0.03 & \ccg{0.02} & -0.02 & \ccr{-0.04} \\
\hline
\multicolumn{1}{l|}{} &  & \multicolumn{2}{c}{MS@k} & \multicolumn{2}{c}{mS@k} & \multicolumn{2}{c}{MS@k} & \multicolumn{2}{c}{mS@k} \\
\multicolumn{1}{l|}{} &  & $[A]$ & $[A]_D$ & $[A]$ & $[A]_D$ & $[B]$ & $[B]_D$ & $[B]$ & $[B]_D$ \\
\hline
\multirow{2}{*}{Race} & pos & 0.61 & \ccg{0.50} & -1.17 & \ccg{-1.06} & 0.61 & \ccg{0.51} & -0.49 & \ccr{-1.21} \\
 & neg & 0.65 & \ccg{0.59} & -1.19 & \ccg{-1.18} & 0.63 & \ccg{0.51} & -0.88 & \ccr{-1.01} \\
\multirow{2}{*}{Gender} & pos & 0.24 & \ccg{0.16} & -0.38 & \ccg{-0.23} & 0.19 & \ccg{0.11} & -0.31 & \ccg{-0.12} \\
 & neg & 0.33 & \ccg{0.24} & -0.64 & \ccg{-0.41} & 0.19 & \ccg{0.18} & -0.29 & \ccg{-0.20} \\
\multirow{2}{*}{Age} & pos & 0.42 & \ccr{0.43} & -0.22 & \ccr{-0.26} & 0.35 & \ccg{0.23} & -0.22 & \ccr{-0.63} \\
 & neg & 0.49 & \ccg{0.31} & -0.53 & \ccg{-0.29} & 0.41 & \ccg{0.26} & -0.34 & \ccr{-0.92}\\
 \bottomrule
\end{tabular}
\end{table}

% Please add the following required packages to your document preamble:
% \usepackage{multirow}
\begin{table}[]
\caption{The Max-/Min-Skew scores for the \pata dataset for the different variants of ViT-based CLIP. $[B_{s}]$ is for CLIP-ViT-B/16, and $[L]$ is for ViT-L/14.}
\label{app:tab:vits}
\vspace{-0.5em}
\setlength{\tabcolsep}{2pt}
\renewcommand{\arraystretch}{0.9}
\begin{tabular}{c|l|llllllll}
\toprule
\multicolumn{1}{l|}{PA} & +/- & \multicolumn{2}{c}{MSkew} & \multicolumn{2}{c}{mSkew} & \multicolumn{2}{c}{MSkew} & \multicolumn{2}{c}{mSkew} \\
\multicolumn{1}{l|}{} &  & $B_{s}$ & $[B_{s}]_D$ & $[B_{s}]$ & $[B_{s}]_D$ & $[L]$ & $[L]_D$ &  $[L]$ & $[L]_D$ \\
\midrule
\multirow{2}{*}{Race} & pos & 0.03 & 0.03 & -0.04 & -0.04 & 0.25 & 0.25 & -0.54 & \ccg{-0.52} \\
 & neg & 0.04 & \ccg{0.03} & -0.04 & -0.04 & 0.28 & \ccg{0.27} & -0.46 & -0.46 \\
\multirow{2}{*}{Gender} & pos & 0.01 & 0.01 & -0.01 & -0.01 & 0.12 & \ccg{0.09} & -0.16 & \ccg{-0.11} \\
 & neg & 0.02 & \ccg{0.01} & -0.02 & \ccg{-0.01} & 0.14 & \ccg{0.12} & -0.20 & \ccg{-0.18} \\
\multirow{2}{*}{Age} & pos & 0.01 & 0.01 & -0.01 & -0.01 & 0.16 & \ccr{0.18} & -0.23 & \ccr{-0.27} \\
 & neg & 0.01 & \ccr{0.02} & -0.02 & -0.02 & 0.23 & \ccr{0.29} & -0.36 & \ccr{-0.48}\\
 \bottomrule
\end{tabular}
\end{table}

\subsection{Zero-shot Results}
In Table~\ref{apptab:zeroshot}, we present our zero-shot evaluation for the debiased VLM networks, as described in Section~\ref{sec:evaluation} of the paper. Table~\ref{app:video} presents zero-shot evaluation for the debiased VLMs for video datasets.  
\begin{table}%[t]
\small
\setlength{\tabcolsep}{2.1pt}
\renewcommand{\arraystretch}{0.9}
\centering
\caption{Results of state-of-the-art visual-language models and their \method counterparts for four image classification datasets. Across seven pre-trained visual-language models, \method achieves zero-shot performance similar to vanilla models.
}
\label{apptab:zeroshot}
{
% \resizebox{8.3cm}{!}{
\begin{tabular}{lcccc}
    \toprule
    Model & C-10 & C-100 & FER2013 & ImageNet\\
    \midrule
    \midrule
    CLIP \begin{scriptsize}(ViT/B-32)\end{scriptsize} & 89.93 & 62.93 & 43.83 & 58.08\\
    \method-CLIP \begin{scriptsize}(ViT/B-32)\end{scriptsize} & 88.85 & 60.08 & 39.60 & 55.84\\
    $\Delta$ & 1.08 & 2.85 & 4.23 & 2.24\\
    \midrule
    CLIP \begin{scriptsize}(ViT/B-16)\end{scriptsize} & 90.96 & 67.49 & 50.74 & 63.64\\
    \method-CLIP \begin{scriptsize}(ViT/B-16)\end{scriptsize} & 90.23 & 66.16 & 49.33 & 61.36\\
    $\Delta$ & 0.73 & 1.33 & 1.41 & 2.28\\
    \midrule
    CLIP \begin{scriptsize}(ViT/L-14)\end{scriptsize} & 95.73 & 76.64 & 46.16 & 71.22\\
    \method-CLIP \begin{scriptsize}(ViT/L-14)\end{scriptsize} & 95.26 & 75.68 & 42.33 & 66.43\\
    $\Delta$ & 0.47 & 0.96 & 3.83 & 2.24\\
    \midrule
    CLIP \begin{scriptsize}(RN50)\end{scriptsize} & 74.06 & 40.89 & 37.67 & 55.22\\
    \method-CLIP \begin{scriptsize}(RN50)\end{scriptsize} & 72.36 & 39.73 & 40.95 & 52.96\\
    $\Delta$ & 1.7 & 1.16 & -3.28 & 2.26\\
    \midrule
    FLAVA & 90.53 & 65.60 & 28.36 & 49.30 \\
    \method-FLAVA & 89.05 & 64.00 & 27.19 & 47.67 \\
    $\Delta$ & 1.48 & 1.60 & 1.17 & 1.63\\
    \midrule
    BLIP & 85.00 & 51.61 & 39.50 & 32.57 \\
    \method-BLIP & 81.20 & 48.90 & 36.50 & 29.94 \\
    $\Delta$ & 3.80 & 2.71 & 3.00 & 2.63\\
    \midrule
    ALBEF & 84.00 & 50.61 & 39.39 & 31.57 \\
    \method-ALBEF & 80.20 & 47.80 & 35.89 & 29.92 \\
    $\Delta$ & 3.80 & 2.81 & 3.50 & 1.65\\
    \bottomrule
\end{tabular}}
% }
\end{table}

\begin{table}[!ht]
\small
\setlength{\tabcolsep}{1.5pt}
\renewcommand{\arraystretch}{0.9}
\centering
\caption{Results of state-of-the-art visual-language models and their \method counterparts for three video classification datasets. Across five pre-trained visual-language models, \method achieves zero-shot performance similar to vanilla models.
}
\label{app:video}
{
% \resizebox{8.3cm}{!}{
\begin{tabular}{lcccc}
    \toprule
    \multirow{2}{*}{Model} & UCF-101 & Kinetics-700 & \multicolumn{2}{c}{RareAct}\\
    & Top-1 & AVG & mWAP & mWSAP\\
    \midrule
    CLIP \begin{scriptsize}(ViT/B-32)\end{scriptsize} & 57.65 & 43.97 & 16.63 & 16.78\\
    \method-CLIP \begin{scriptsize}(ViT/B-32)\end{scriptsize} & 55.77 & 42.20 & 16.02 & 16.03\\
    $\Delta$ & 1.88 & 1.77 & 0.61 & 0.75 \\
    \midrule
    CLIP \begin{scriptsize}(ViT/B-16)\end{scriptsize} & 59.55 & 48.38 & 18.58 & 18.69\\
    \method-CLIP \begin{scriptsize}(ViT/B-16)\end{scriptsize} & 56.53 & 46.49 & 17.54 & 17.66\\
    $\Delta$ & 3.02 & 1.89 & 1.04 & 1.03 \\
    \midrule
    CLIP \begin{scriptsize}(ViT/L-14)\end{scriptsize} & 67.88 & 55.86 & 25.42 & 25.55\\
    \method-CLIP \begin{scriptsize}(ViT/L-14)\end{scriptsize} & 67.43 & 53.21 & 25.20 & 25.34\\
    $\Delta$ & 0.45 & 2.65 & 0.22 & 0.21 \\
    \midrule
    CLIP \begin{scriptsize}(RN50)\end{scriptsize} & 52.73 & 39.39 & 15.08 & 15.09\\
    \method-CLIP \begin{scriptsize}(RN50)\end{scriptsize} & 50.25 & 38.59 & 14.41 & 14.54\\
    $\Delta$ & 2.48 & 0.8 & 0.67 & 0.55 \\
    \midrule
    FLAVA & 39.09 & 37.85 & 16.12 & 16.14\\
    \method-FLAVA & 37.27 & 35.59 & 15.30 & 15.43\\
    $\Delta$ & 1.82 & 2.26 & 0.82 & 0.71 \\
    \midrule
    BLIP & 43.26 & 37.07 & 16.35 & 16.44\\
    \method-BLIP & 40.34 & 34.78 & 15.86 & 15.92\\
    $\Delta$ & 2.92 & 2.29 & 0.49 & 0.52 \\
    \midrule
    ALBEF & 22.07 & 26.10 & 15.23 & 15.49\\
    \method-ALBEF & 20.77 & 24.33 & 14.33 & 14.56\\
    $\Delta$ & 1.3 & 1.77 & 0.9 & 0.93 \\
    \bottomrule
\end{tabular}}
% }
\end{table}

% Please add the following required packages to your document preamble:
% \usepackage{multirow}
\subsection{Qualitative Results}
\begin{figure*}[!ht]
    \centering
    \includegraphics[width=\linewidth]{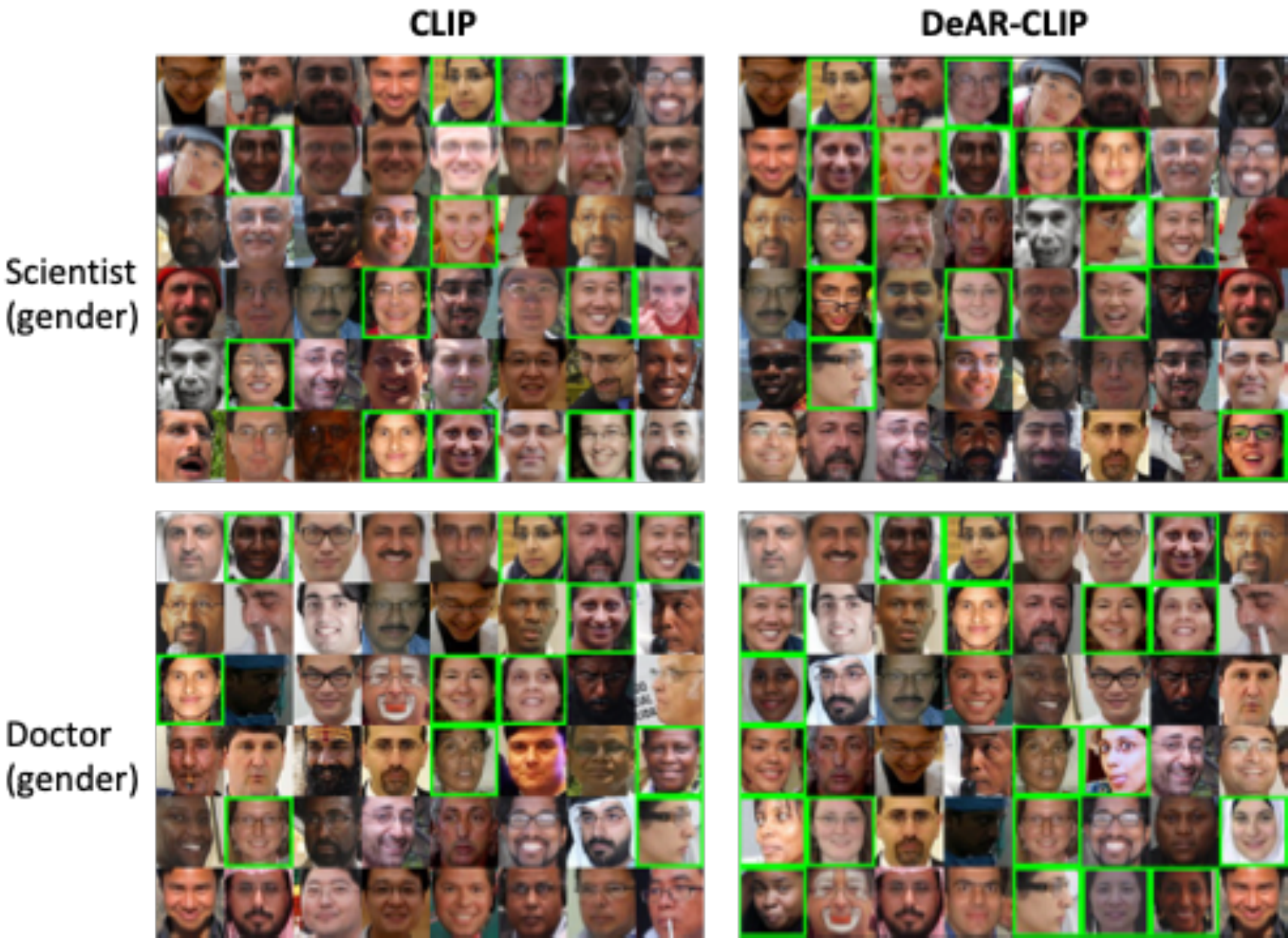}
    \caption{Qualitative comparison on top-k retrieval (k=48) for CLIP (left) and \method-CLIP features (right).}
    \label{fig:qualitative}
\end{figure*}

We also present qualitative results for face image retrieval with text queries (CLIP text features), using the image features generated using CLIP and \method-CLIP. Figure \ref{fig:qualitative} shows a few instances of two phrases. Our results indicate an improvement in the diversity of results. For instance, for the phrases ``photo of a doctor" and ``photo of a  scientist", we see a clear improvement in the gender parity of the returned faces. We note some overlap between the results but the ranks assigned to them are different. Also, we note that the overlap is higher for phrases containing the keyword ``person", and we find that this is so because some images have a much higher text association with the keyword than others.

\section{Further Ablation Studies}
\label{app:ablation}
\vspace{-0.5em}
We present results for further ablation studies as evidence for the effectiveness of the \method framework.

\subsection{Disentanglement of Protected Attributes in ARL residual representation} 
\begin{figure*}[t]
    \centering
    \includegraphics[width=0.9\linewidth]{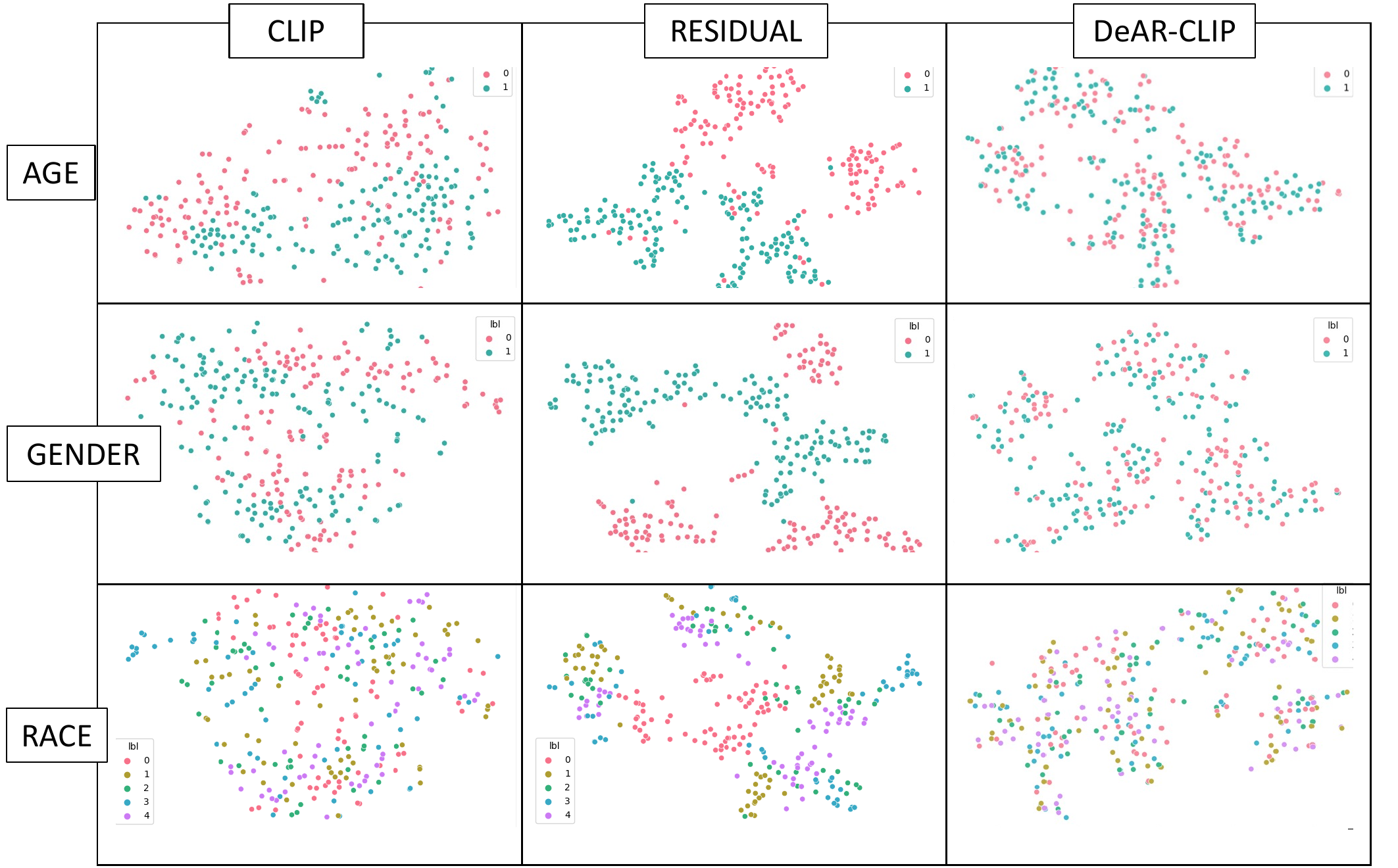}
    \caption{High-resolution version of Figure \ref{fig:tsne} in the paper. TSNE Plots for CLIP, Residual and \method-CLIP features for a subset of the \pata dataset indicate that the residual plots indeed capture the specific attributes and that the \method-CLIP features have greater overlap between points of different protected labels than the original features.}
    \label{app:tsne}
\end{figure*}

Figure \ref{app:tsne} illustrates the degree of disentanglement that the ARL module imposes on CLIP features. The gender and age clusters are distinctly visible (column 2 of the figure), while the ethnic-racial clusters have a slightly worse disentanglement. We attribute that to the lower accuracy of the race classifier (PAC) trained using CLIP features on the FairFace dataset. We also observe that after adding the residual, point co-incidences increase considerably over the base model's plot, indicating that the \method-CLIP model is worse at identifying gender, race and age than the vanilla CLIP model. 

\subsection{Joint-training for PAC and ARL in an adversarial setting} Previous approaches like Berg et al.~\cite{berg2022prompt} use adversarial training of a protected-attribute classifier (PAC). We attempt to use the same approach with our ARL model and find much worse performance on the Max-Skew and Min-skew scores. This is because the network does not converge (even with modified hyperparameters) to the joint minimum for the classifier losses ($L_{\text{ce}}$) and the reconstruction loss ($L_{\text{recon}}$).
\begin{table}[!h]
\centering
\caption{Ablation results for joint training of the PAC and ARL modules The joint training does not yield the expected de-biasing effect because the network does not converge to a common minimum between the $L_{\text{recon}}$ and $-L_{\text{ce}}$}
\vspace{-0.5em}
\label{app:pataskew}
\setlength{\tabcolsep}{2pt}
\renewcommand{\arraystretch}{1}
\begin{tabular}{c|c|cccccccc}
\toprule
\multirow{2}{*}{PA} & \multirow{2}{*}{+/-} & \multicolumn{2}{c}{MaxSkew} & \multicolumn{2}{c}{MinSkew}  \\
 && \multicolumn{1}{c}{[C]} & \multicolumn{1}{c}{[C]$_{\text{D}}$} & \multicolumn{1}{c}{[C]} & \multicolumn{1}{c}{[C]$_{\text{D}}$} \\
 \hline
\multirow{2}{*}{Age} & +ve & 0.10 & \ccr{0.23} & -0.12 & \ccr{-0.31}   \\
 & -ve & 0.19 & 0.19 & -1.21 & \ccg{-0.27} \\
 \multirow{2}{*}{Race} & +ve & 0.16 & \ccr{0.39} & -0.43 & \ccr{-1.22} \\
 & -ve  & 0.45 & \ccg{0.41} & -3.40 & \ccg{-3.21}  \\
 \multirow{2}{*}{Gender} & +ve & 0.09 & \ccr{0.18} & -0.11 & \ccr{-0.27} \\
 & -ve  & 0.21 & \ccg{0.19} & -0.79 & \ccg{-0.73}  \\
\bottomrule
\end{tabular}
\end{table}

\begin{figure*}
    \includegraphics[width=\textwidth]{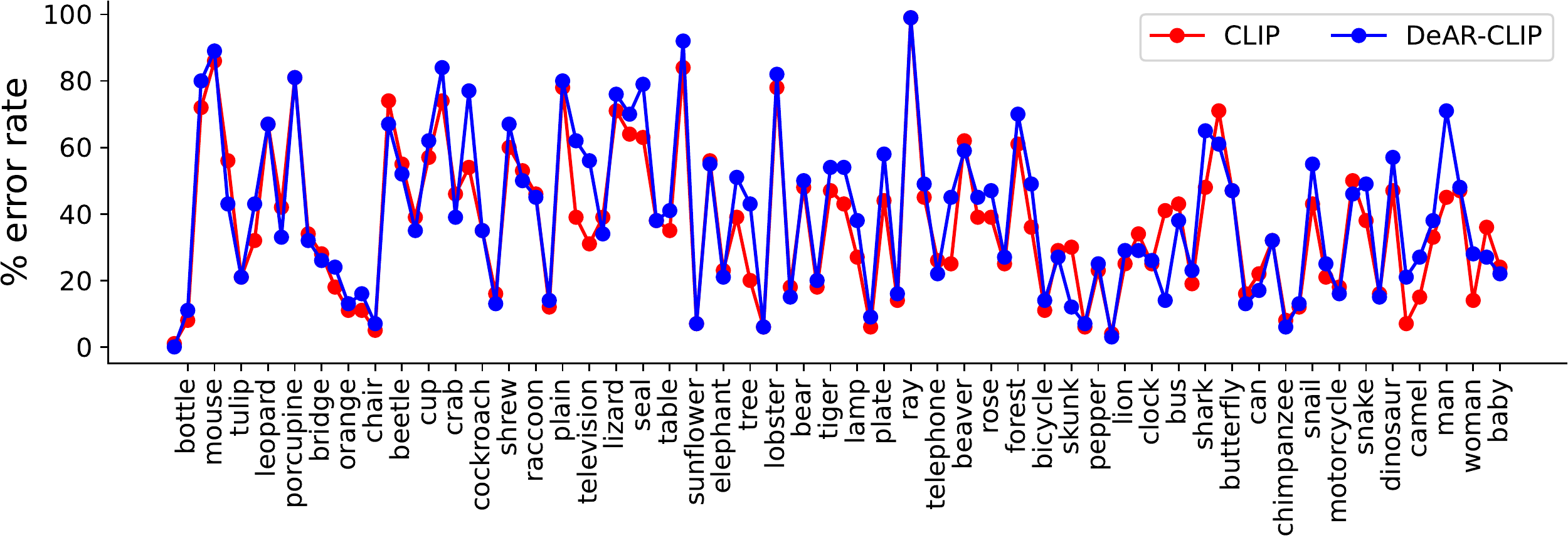}
    \caption{Comparing class error rate of CIFAR-100 for vanilla CLIP Vs \method-CLIP. We observe an increase in error rate for only human-related labels, \eg, an increase from 45\% to 71\% in the error rate of the ``man'' class after debiasing.}
\end{figure*}

\subsection{Error analysis for zero-shot tasks} We compare the class error rate of CIFAR-100 for CLIP and \method-CLIP. We observe an increase in error rate for only human-related labels, \eg, an increase from 45\% to 71\% in the error rate of the \emph{man} class after debiasing. This proves that the debiasing framework is successful at paying more attention to the features that characterize protected attributes such as \texttt{gender}, aligning with the overall objective of \method. 

\subsection{Extending \method for unimodality} Next, we extend our proposed \method framework to unimodal models, where we take image representations from unimodal ViT/B-16, and ViT/B-14 models pre-trained on ImageNet and then train a linear layer on top of it. For CIFAR-10 and CIFAR-100, we observe a classification accuracy drop of \textbf{1.1\%} and \textbf{0.8\%}, respectively. Further, we observe that debiasing leads to a uniform accuracy drop across all protected attributes, \ie, it decomposes the visual representation so that the protected information is subtracted out. 

\section{Limitations and Future Work}
Our work presents the first step towards debiasing VLMs and as such, we observe its limitations in several respects:
\begin{enumerate}
\item The association of sub-string matches, such as the text ``person" causes keywords with the \textit{person} suffix or phrases with the keyword in it to behave differently than expected. For instance, the keyword business-person has a different association (as measured by the max-skew) than the keyword ``business". This causes overall skew distributions to be inaccurate, and incommensurate with the qualitative assessment.
\item We also observe that the network often over-compensates flipping the skew in favor (or disfavor) of a different protected label. For instance, in the case of ALBEF~\cite{ALBEF}, using the \method framework (Table~\ref{app:tab:pataskew}), we observe that the skew increases for the Age-Positive combination. However, we also find that it flips over from being in favor of ``young" people to that of ``old" people. We attribute this flipping behavior to the inaccuracy of training of the Age-classifier in the PAC module, and we look to improve its accuracy of it by modifying its hyperparameters or architectures.
\item We also observe a slight increase in skew values for the FLAVA model using \method framework for Race-Postive and Race-Negative combinations. We attribute this to the inaccuracy of PAC in classifying race. We hypothesize alleviating this behavior by modifying the training hyperparameters or architectures of \method.
\item We recognize that the skew analysis is highly sensitive to its parameters (thresholds, the value of k, and choice of text prompts), and we look to address these with uniform metrics in the future. 
\item The first version of our proposed \pata dataset does not cover the entire ground to determine the fairness of a VLM. We look to expand the categories set to include more scenes and queries.
\item The \method framework appears not to work very well for all variants of the CLIP network. (Table \ref{app:tab:vits}). We attribute this again to the inaccurate PAC module. 
\end{enumerate}

% \paragraph{Paper Errata}
% We found one scene category in the \pata dataset to have no images of the Hispanic ethnic-racial group. This does not affect the analysis significantly. We will rectify this in the final version of the dataset.

\end{document}